\documentclass[journal]{IEEEtai}
\usepackage[colorlinks,urlcolor=blue,linkcolor=blue,citecolor=blue]{hyperref}
\usepackage{times}
\usepackage{epsfig}
\usepackage{graphicx}
\usepackage{amsmath}
\usepackage{amssymb}
\usepackage{commath}
\usepackage{amsthm}
\usepackage{algorithmic}
\usepackage[super]{nth}
\usepackage[norelsize,longend,ruled,lined,boxed,commentsnumbered]{algorithm2e}
\usepackage{breqn}
\usepackage{multirow}
\newtheorem{theorem}{Theorem}
\newtheorem{corollary}{Corollary}

\begin{document}

\title{AdaInject: Injection Based Adaptive Gradient Descent Optimizers for Convolutional Neural Networks}

\author{
Shiv Ram Dubey, \IEEEmembership{Senior Member,~IEEE}, S.H. Shabbeer Basha, Satish Kumar Singh, \IEEEmembership{Senior Member,~IEEE}, and Bidyut Baran Chaudhuri, \IEEEmembership{Life Fellow,~IEEE}
\thanks{S.R. Dubey and S.K. Singh are with the Computer Vision and Biometrics Laboratory (CVBL), Indian Institute of Information Technology, Allahabad, Prayagraj, Uttar Pradesh-211015, India (e-mail: srdubey@iiita.ac.in, sk.singh@iiita.ac.in). }
\thanks{S.H.S. Basha is with the PathPartner Technology Pvt. Ltd., Bangalore, India (e-mail: shabbeer.sh@pathpartnertech.com). }
\thanks{B.B. Chaudhuri was with the Indian Statistical Institute, Kolkata, India and now associated with Techno India University, Kolkata, India (e-mail: bidyutbaranchaudhuri@gmail.com). }
\thanks{\textbf{This paper is accepted for publication by IEEE Transactions on Artificial Intelligence.}
\textbf{Copyright © 2022 IEEE. Personal use of this material is permitted. However, permission to use this material for any other purposes must be obtained from the IEEE by sending an email to pubs-permissions@ieee.org.}}
}
\markboth{IEEE Transactions on Artificial Intelligence}
 {AdaInject Optimizer}

\maketitle

\begin{abstract}
The convolutional neural networks (CNNs) are generally trained using stochastic gradient descent (SGD) based optimization techniques. The existing SGD optimizers generally suffer with the overshooting of the  minimum and oscillation near minimum. In this paper, we propose a new approach, hereafter referred as AdaInject, for the gradient descent optimizers by injecting the second order moment into the first order moment. Specifically, the short-term change in parameter is used as a weight to inject the second order moment in the update rule. The AdaInject optimizer controls the parameter update, avoids the overshooting of the minimum and reduces the oscillation near minimum. The proposed approach is generic in nature and can be integrated with any existing SGD optimizer. The effectiveness of the AdaInject optimizer is explained intuitively as well as through some toy examples. We also show the convergence property of the proposed injection based optimizer. Further, we depict the efficacy of the AdaInject approach through extensive experiments in conjunction with the state-of-the-art optimizers, namely AdamInject, diffGradInject, RadamInject, and AdaBeliefInject on four benchmark datasets. Different CNN models are used in the experiments. A highest improvement in the top-1 classification error rate of $16.54\%$ is observed using diffGradInject optimizer with ResNeXt29 model over the CIFAR10 dataset. Overall, we observe very promising performance improvement of existing optimizers with the proposed AdaInject approach. The code is available at: \url{https://github.com/shivram1987/AdaInject}. 
\end{abstract}

\begin{IEEEImpStatement}
Adaptive moment based optimizers are among the popular gradient descent optimization techniques for the training of deep learning models. They try to control the step size based on the gradient behavior. However, the existing gradient descent optimization techniques either overshoot the ``steep and narrow'' valley (i.e., minimum) or oscillate near it, due to large step size caused by the exponential moving average of gradients used for parameter updates. The AdaInject optimization technique we introduce in this paper tackled this problem by incorporating the immediate parameter change weighted second order moment injection for the parameter updates. 
Using the proposed optimization technique, a significant improvement is observed in the performance of image classification using different CNN models. Moreover, the proposed AdaInject approach can be used with any existing adaptive moment based optimization technique. Hence, it can provide the alternative optimizers with better step size control to train different deep learning models for diverse applications.
\end{IEEEImpStatement}


\section{Introduction}
\IEEEPARstart{D}{eep} learning has shown a great impact over the performance of the neural networks for a wide range of problems \cite{deeplearning}. In recent past, convolutional neural networks (CNNs) have shown very promising results for different computer vision applications, such as object recognition \cite{ResNet}, \cite{resnext}, \cite{densenet}, \cite{kantipudi2020color}; object detection \cite{fastrcnn}, \cite{fasterrcnn}; face recognition \cite{facenet}, \cite{deepface}; image quality assessment \cite{pan2022no}; gesture recognition \cite{zou2021transfer}; Covid-19 grading \cite{de2021automated}; and many more. CNNs have also been used as basic building blocks in other networks like Autoencoder \cite{zeng2015coupled}, \cite{chen2019bae}, \cite{dewangan2021fault}, Siamese Network \cite{dong2018triplet}, \cite{fan2019siamese}, Generative Adversarial Networks \cite{zhu2017unpaired}, \cite{babu2020pcsgan}, etc.

The training of different types of deep neural networks (DNNs) is mainly performed with the help of stochastic gradient descent (SGD) based optimization \cite{sgd1}. SGD optimizer updates the parameters of the network based on the gradient of objective function w.r.t. the corresponding parameters \cite{sgd}. The vanilla SGD optimization suffers from three problems, including 1) zero gradient in local minimum and saddle regions leading to no update in the parameters, 2) a jittering effect along steep dimensions due to the inconsistent changes in the loss caused by the different parameters, and 3) noisy updates due to the gradient computed from the batch of data. SGD with moment (SGDM) \cite{sgdm} considers the first order moment (i.e., velocity) as an exponential moving average (EMA) of gradient for each parameter while training progresses \cite{SGDM1}. The parameter is updated in SGDM based on the EMA of gradient which resolves the problem of zero gradient.

Several SGD based optimization techniques have been proposed in the recent past \cite{AdaGrad}, \cite{AdaDelta}, \cite{RMSProp}, \cite{Adam}, \cite{diffgrad}, \cite{radam}, \cite{adabelief}, and etc. AdaGrad \cite{AdaGrad} controls the learning rate by dividing it with the root of the sum of the squares of the past gradients. However, it makes the learning rate very small after certain iterations and kills the parameter update. AdaDelta \cite{AdaDelta} resolves the diminishing learning rate issue of AdaGrad by considering only a few immediate past gradients. However, it is not able to exploit the global information. In another attempt to resolve the problem of AdaGrad, RMSProp \cite{RMSProp} divides the learning rate by the root of the exponentially decaying average of squared gradients. In 2015, Kingma and Ba \cite{Adam} proposed the adaptive moment based Adam optimizer. Adam combines the ideas of SGDM and RMSprop and uses first order and second order moments. Adam computes the first order moment as the EMA of gradients and uses it to update the parameter. Adam also computes the second order moment as the EMA of the square of gradients and uses it to control the learning rate. Adam performs well in practice to train the convolutional neural networks (CNNs) \cite{Adam}. However, it suffers from overshooting and oscillations near minimum and varying gradient variance due to batch wise computation. diffGrad \cite{diffgrad} resolved the issues as posed by Adam by introducing a friction term in parameter update using the rate of change in gradients. Radam \cite{radam} resolved the variance issue as posed by Adam by rectifying the variance of gradients during parameter update. AdaBelief \cite{adabelief} uses the belief in gradients to compute the second order moment. The belief in gradients is computed as the difference between the gradient and the first order moment of the gradient. Other recently proposed and notable gradient descent optimizers are Proportional Integral Derivative (PID) \cite{pid}, Nesterov’s Moment Adam (NADAM) \cite{NADAM}, Nostalgic Adam (NosAdam) \cite{nosadam}, YOGI \cite{yogi}, Adaptive Bound (AdaBound) \cite{adabound}, Adaptive and Momental Bound (AdaMod) \cite{adamod}, Aggregated Moment (AggMo) \cite{aggmo}, Lamb \cite{lamb}, Adam Projection (AdamP) \cite{adamp}, Gradient Centralization (GC) \cite{gc}, AdaHessian \cite{adahessian}, and AngularGrad \cite{roy2021angulargrad}.

\begin{algorithm}[!t]
\caption{Adam Optimizer}
\SetAlgoLined
\textbf{Initialize:} $\theta_{0},m_{0}\gets0,v_{0}\gets0,t\gets0$\\
\textbf{Hyperparameters:} $\alpha, \beta_1, \beta_2$\\
\textbf{While} $\theta_{t}$ not converged\\
    \hspace{0.45cm} $t \gets t+1$\\
    \hspace{0.45cm} $g_t \gets \nabla_{\theta} f_t(\theta_{t-1})$ \\
    \hspace{0.45cm} $m_t \gets \beta_1 \cdot m_{t-1} + (1-\beta_1) \cdot g_t$\\
    \hspace{0.45cm} $v_t \gets \beta_2 \cdot v_{t-1} + (1-\beta_2) \cdot g^2_t$\\
    \hspace{0.45cm} \textbf{Bias Correction}\\
    \hspace{0.9cm} $\widehat{m_t} \gets m_t/(1-\beta_1^t)$, $\widehat{v_t} \gets v_t/(1-\beta_2^t)$\\
    \hspace{0.45cm} \textbf{Update}\\
    \hspace{0.9cm} $\theta_t \gets \theta_{t-1} - \alpha \cdot \widehat{m_t}/(\sqrt{\widehat{v_t}} + \epsilon)$
\end{algorithm}

The adaptive SGD optimization techniques have led to a promising performance on deep CNN models. 
The majority of the above mentioned adaptive gradient descent optimizers suffer due to the overshooting of the minimum and oscillation near minimum. However, it is evident that a robust online stepsize adaptation in optimization plays an important role in gradient descent optimization \cite{curvature}. 
We resolve the above issues by injecting the second order moment in first order for the parameter update, which is weighted by the short-term parameter update history to incorporate the robust adaptation of step size.
The major contribution of this work is summarized as follows:
\begin{itemize}
    \item We propose AdaInject for the adaptive optimizers by injecting the short-term parameter change weighted second order moment in EMA of gradient used for parameter update.
    \item We provide an intuitive explanation in support of the effectiveness of the proposed AdaInject in different optimization scenarios.
    \item We show the effect of the proposed approach using toy examples. The convergence analysis is also conducted using regret bound which shows the convergence property of the proposed approach.
    \item We validate the superiority of the proposed injection concept with the recent state-of-the-art optimizers, including Adam \cite{Adam}, diffGrad \cite{diffgrad}, Radam \cite{radam} and AdaBelief \cite{adabelief} using a wide range of CNN models for image classification over four benchmark datasets. 
    \item The proposed concept is generic and can be easily integrated with any existing adaptive moment based SGD optimizer.
\end{itemize}

\begin{algorithm}[!t]
\caption{AdamInject (i.e., Adam + AdaInject) Optimizer}
\SetAlgoLined
\textbf{Initialize:} $\theta_{0},s_{0}\gets0,v_{0}\gets0,t\gets0$\\
\textbf{Hyperparameters:} $\alpha, \beta_1, \beta_2, k$\\
\textbf{While} $\theta_{t}$ not converged\\
    \hspace{0.45cm} $t \gets t+1$\\
    \hspace{0.45cm} $g_t \gets \nabla_{\theta} f_t(\theta_{t-1})$ \\
    \hspace{0.45cm} \textbf{If} t = 1 \\
    \hspace{0.9cm} $s_t \gets \beta_1 \cdot s_{t-1} + (1-\beta_1) \cdot g_t$\\
    \hspace{0.45cm} \textbf{Else} \\
    \hspace{0.9cm} $\textcolor{blue}{\Delta \theta \gets \theta_{t-2} - \theta_{t-1}}$\\
    \hspace{0.9cm} $s_t \gets \beta_1 \cdot s_{t-1} + (1-\beta_1) \cdot \textcolor{blue}{(g_t + \Delta \theta \cdot g^2_t)/k}$\\
    \hspace{0.45cm} $v_t \gets \beta_2 \cdot v_{t-1} + (1-\beta_2) \cdot g^2_t$\\
    \hspace{0.45cm} \textbf{Bias Correction}\\
    \hspace{0.9cm} $\widehat{s_t} \gets s_t/(1-\beta_1^t)$, $\widehat{v_t} \gets v_t/(1-\beta_2^t)$\\
    \hspace{0.45cm} \textbf{Update}\\
    \hspace{0.9cm} $\theta_t \gets \theta_{t-1} - \alpha \widehat{s_t}/(\sqrt{\widehat{v_t}} + \epsilon)$
\end{algorithm}

The remaining paper is structured by presenting the proposed Injection based optimizers in Section 2; Intuitive explanation and empirical analysis in Section 3; Convergence analysis in Section 4; Experimental analysis in Section 5; and Concluding remarks in Section 6.

\section{Proposed Injection based Optimizers}
As per the conventions used in Adam \cite{Adam}, the aim of gradient descent optimization is to minimize the loss function $f(\theta) \in \mathbb{R}$ where $\theta \in \mathbb{R}^d$ is the parameter. The gradient ($g_t$) at $t^{th}$ step is computed as $g_t \gets \nabla_{\theta} f_t(\theta_{t-1})$. 
Adam computes the first order moment ($m_t$) and second order moment ($v_t$) as the exponential moving average (EMA) of $g_t$ and $g_t^2$, respectively, which can be written as, 
\begin{equation}
    m_t = \beta_1 \cdot m_{t-1} + (1 - \beta_1) \cdot g_t
\end{equation}
\begin{equation}
    v_t = \beta_2 \cdot v_{t-1} + (1 - \beta_2) \cdot g_t^2
\end{equation}
where $\beta_1$ and $\beta_2$ are the smoothing hyperparameters, typically set as $\beta_1 = 0.9$ and $\beta_2 = 0.999$. The $g_t^2$ is computed as $g_t \cdot g_t$ as in the Adam. A bias correction is performed as $\widehat{m_t} \gets m_t/(1-\beta_1^t)$, $\widehat{v_t} \gets v_t/(1-\beta_2^t)$ to avoid very large step size in the initial iterations. The parameter update rule in Adam \cite{Adam} is given as,
\begin{equation}
\theta_t \gets \theta_{t-1} - \alpha \widehat{m_t}/(\sqrt{\widehat{v_t}} + \epsilon)
\end{equation}
where $\alpha$ is the learning rate and $\epsilon = 1e^{-8}$ is a small number for numerical stability to avoid division by zero.
A detailed algorithm of Adam optimizer is summarized in Algorithm 1. The first order moment $m_t$ is used to update the parameters in Adam wherein, the second order moment $v_t$ is used to control the learning rate. It can be noticed that Adam mainly relies on the gradients. 

However, the SGDM considers only the momentum to update the parameters as follows:
\begin{equation}
\theta_t \gets \theta_{t-1} - \alpha m_t.
\end{equation}

In order to utilize the parameter update history information during optimization, we propose a novel concept named AdaInject. Basically, we inject the short-term parameter change weighted second order moment into first order moment to compute the injected moment using the EMA of $(g_t + \Delta \theta \cdot g_t^2)/k$ as,
\begin{equation}
    s_t = \beta_1 \cdot s_{t-1} + (1 - \beta_1) \cdot (g_t + \Delta \theta \cdot g_t^2)/k
\end{equation}
where $\Delta \theta = \theta_{t-2} - \theta_{t-1}$ is the short-term change in parameter $\theta$ to utilize the parameter history information and $k$ is an injection controlling hyperparameter, typically set to $2$ in the experiment. 
The injection of parameter history guided second order moment helps the optimizers to perform the smaller updates near minimum (i.e., ``steep and narrow” valley) to avoid the overshooting and oscillation, while reasonably large updates are used in the small curvature regions. This phenomenon is depicted in Fig. \ref{fig:curvature} with a detailed analysis in the next section.
We perform the bias correction of injected moment and second order moment as $\widehat{s_t} \gets s_t/(1-\beta_1^t)$ and $\widehat{v_t} \gets v_t/(1-\beta_2^t)$, respectively. 

The parameter ($\theta$) update of AdamInject optimizer is given as,
\begin{equation}
    \theta_t \gets \theta_{t-1} - \alpha \cdot \widehat{s_t}/(\sqrt{\widehat{v_t}} + \epsilon)
\end{equation}
where $\alpha$ is the learning rate and $\epsilon = 1e^{-8}$ is a small number for numerical stability to avoid the division by zero. We refer to Adam optimizer with the proposed second order moment injection as AdamInject optimizer. A detailed algorithm of AdamInject optimizer is presented in Algorithm 2 with highlighted changes in blue color as compared to vanilla Adam optimizer which is shown in Algorithm 1.

Basically, we use the proposed AdaInject concept with four existing state-of-the-art optimizers, including Adam \cite{Adam}, diffGrad \cite{diffgrad}, Radam \cite{radam} and AdaBelief \cite{adabelief}, and propose the corresponding AdamInject (i.e., Adam + AdaInject), diffGradInject (i.e., diffGrad + AdaInject), RadamInject (i.e., Radam and AdaInject) and AdaBeliefInject (i.e., AdaBelief + AdaInject) optimizers, respectively. The algorithms for different optimizers (i.e., without and with AdaInject), such as diffGrad, diffGradInject, Radam, RadamInject, AdaBelief, and AdaBeliefInject, are provided in Supplementary. Though we test the proposed injection concept with four optimizers, it can be extended to any EMA based gradient descent optimization technique. In the next section, we analyze the property of the proposed approach.

\section{Intuitive Explanation and Empirical Analysis}
In this section, we present an intuitive explanation using a one dimensional optimization landscape having three scenarios and an empirical analysis using three toy examples.

\subsection{Intuitive Explanation}
\begin{figure}[!t]
    \centering
    \includegraphics[width=\columnwidth]{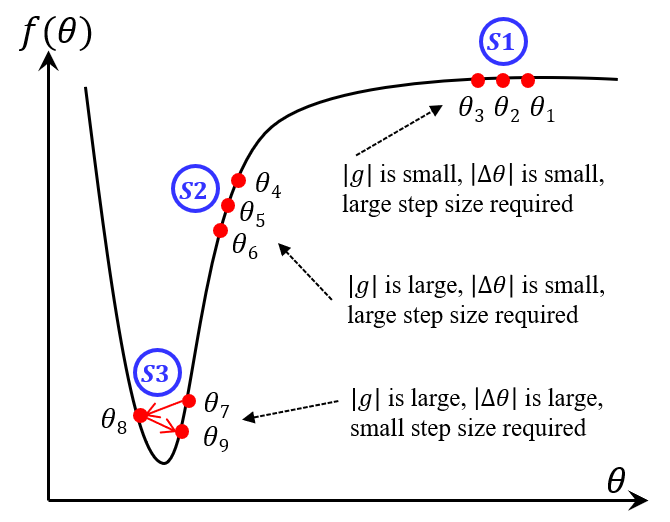}
    \caption{A typical scenario in the optimization depicting the importance of adaptive parameter update in optimization \cite{curvature}, \cite{adabelief}.}
    \label{fig:curvature}
\end{figure}

The existing gradient descent optimizers such as Adam, diffGrad, Radam, etc. only consider the EMA of gradient for parameter update. However, the consideration of parameter history is important as the gradient behavior and required stepsize are different for different regions of loss optimization landscape \cite{curvature}, \cite{adabelief}. We explain the advantange of the proposed optimizer by considering three typical scenarios using a one dimensional optimization curvature (i.e., \textbf{S1}, \textbf{S2} and \textbf{S3}) as depicted in Fig. \ref{fig:curvature}. The bias correction step is ignored in the explanation for simplicity.

\textbf{S1:} This scenario depicts a flat region on the optimization landscape. An ideal optimizer is expected to perform large parameter updates in this scenario. The $|g_t|$ and $|\Delta\theta|$ in flat region are small. Thus, the EMA of gradient (i.e., $m_t$) as well as the EMA of proposed injected gradient (i.e., $s_t$) are small. It leads to a small stepsize in SGD. However, the step size is sufficiently large in both Adam and AdaInject due to the small value of $\sqrt{v_t}$ in the denominator.

\begin{figure*}
    \centering
    \includegraphics[width=0.325\textwidth]{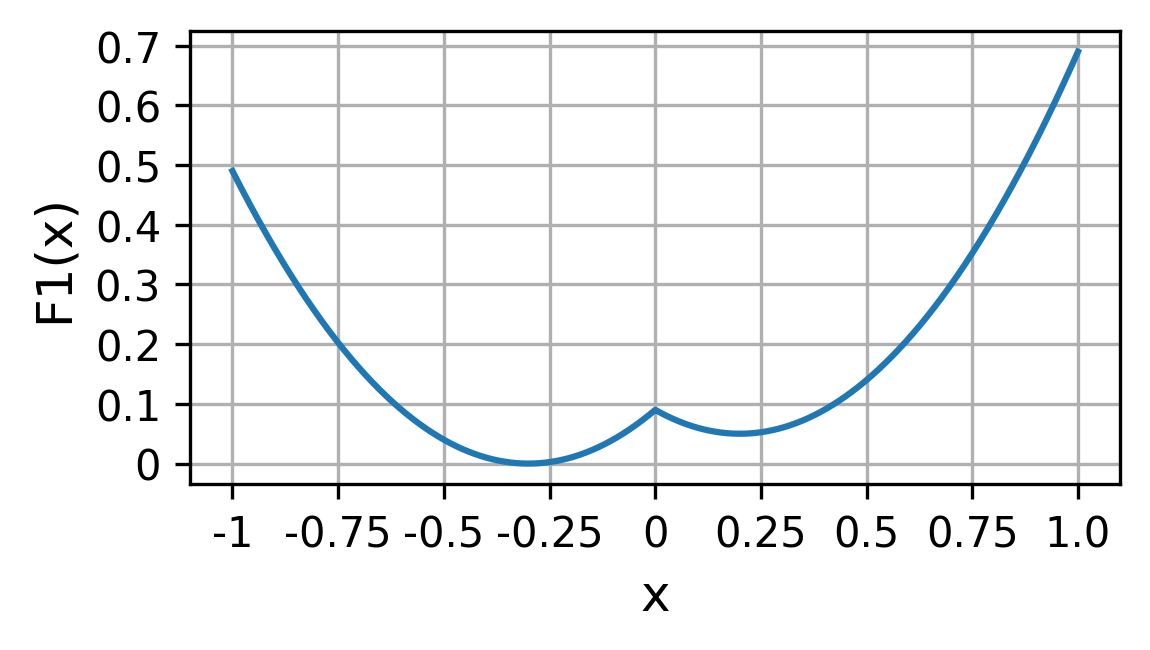}
    \includegraphics[width=0.325\textwidth]{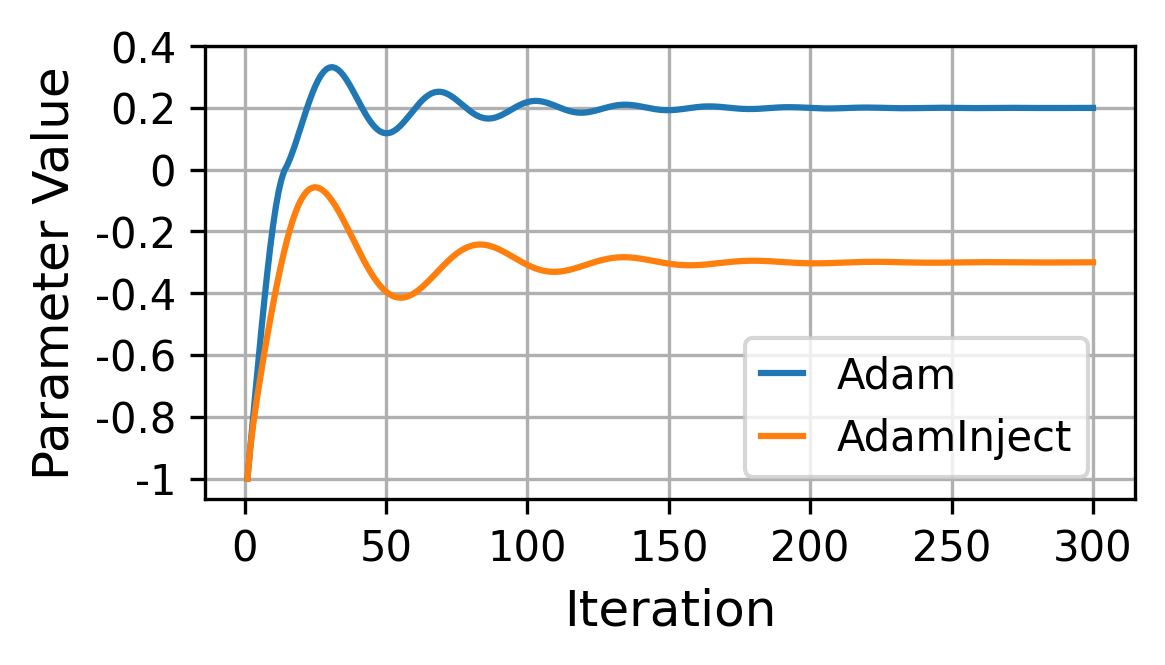}
    \includegraphics[width=0.325\textwidth]{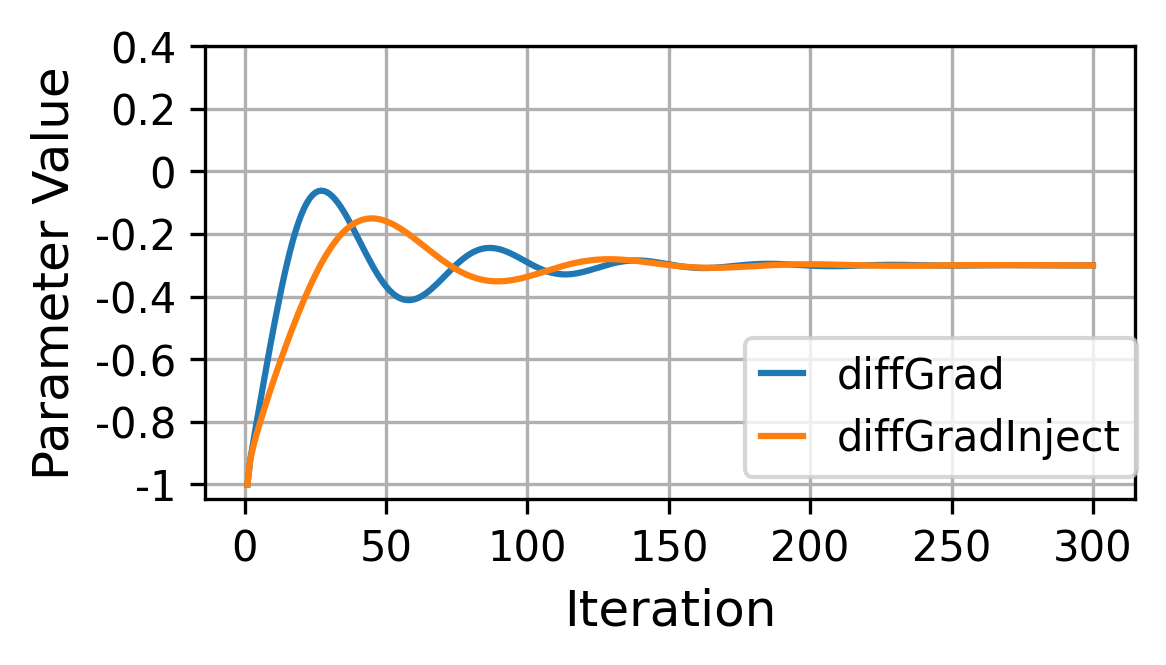} \\
    \includegraphics[width=0.325\textwidth]{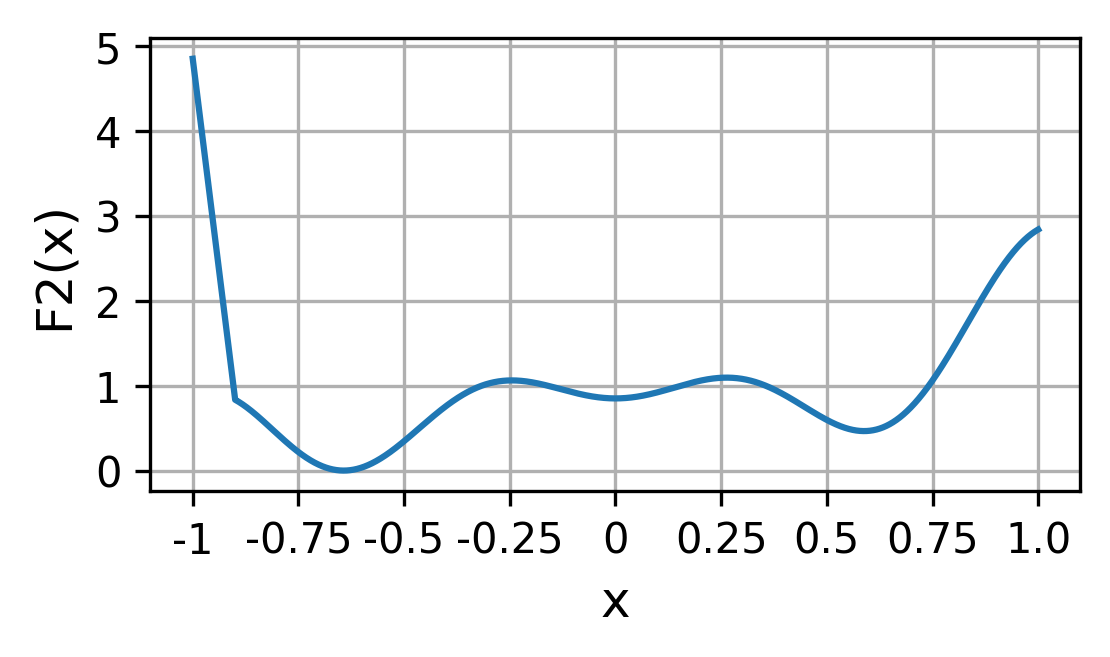}
    \includegraphics[width=0.325\textwidth]{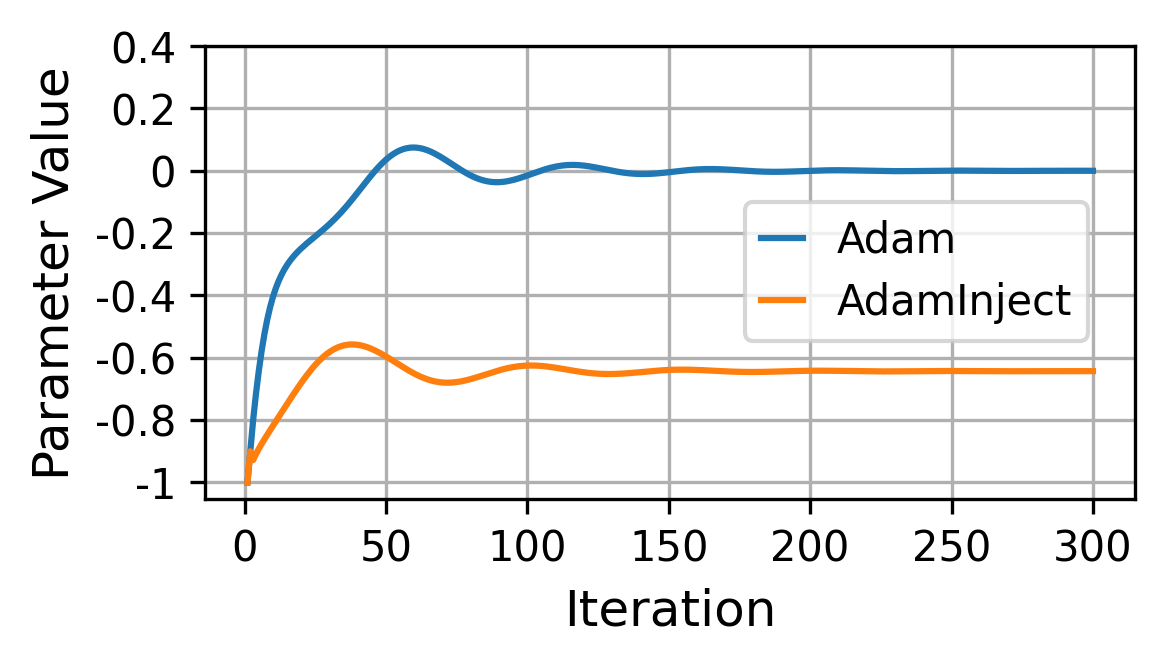}
    \includegraphics[width=0.325\textwidth]{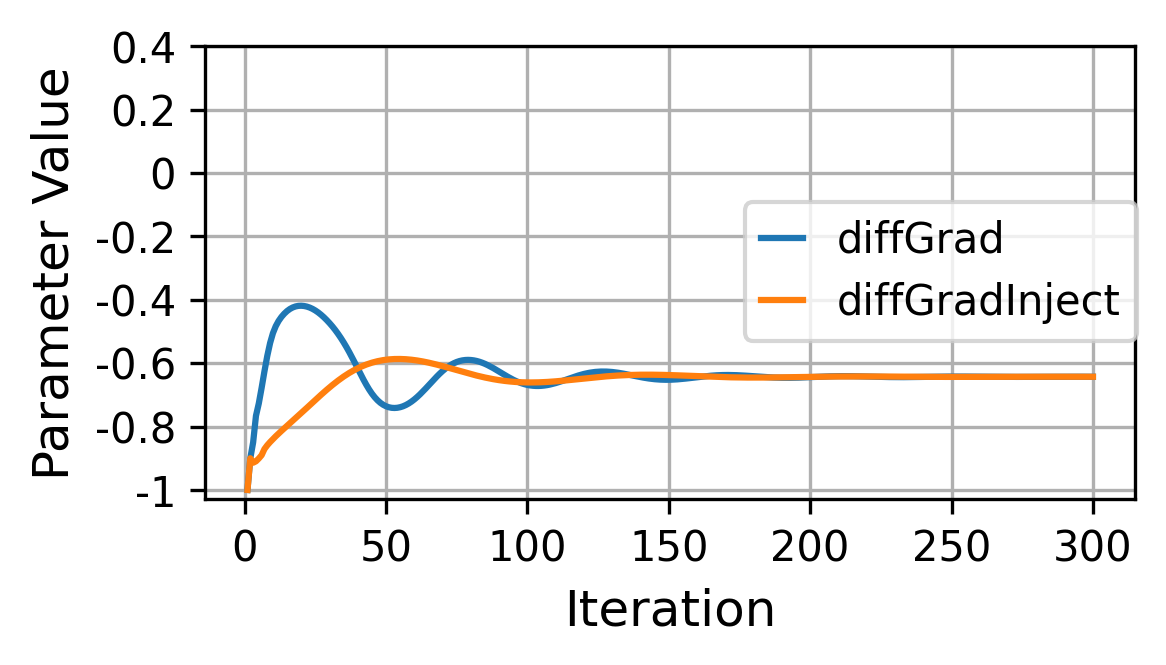}\\
    \includegraphics[width=0.325\textwidth]{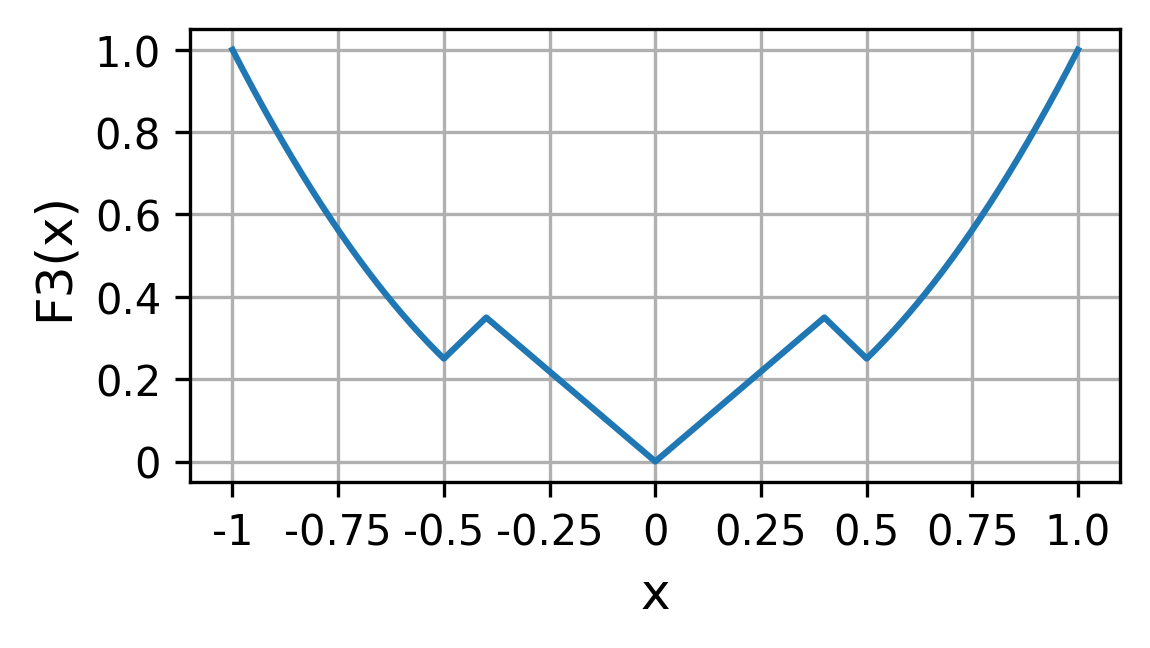}
    \includegraphics[width=0.325\textwidth]{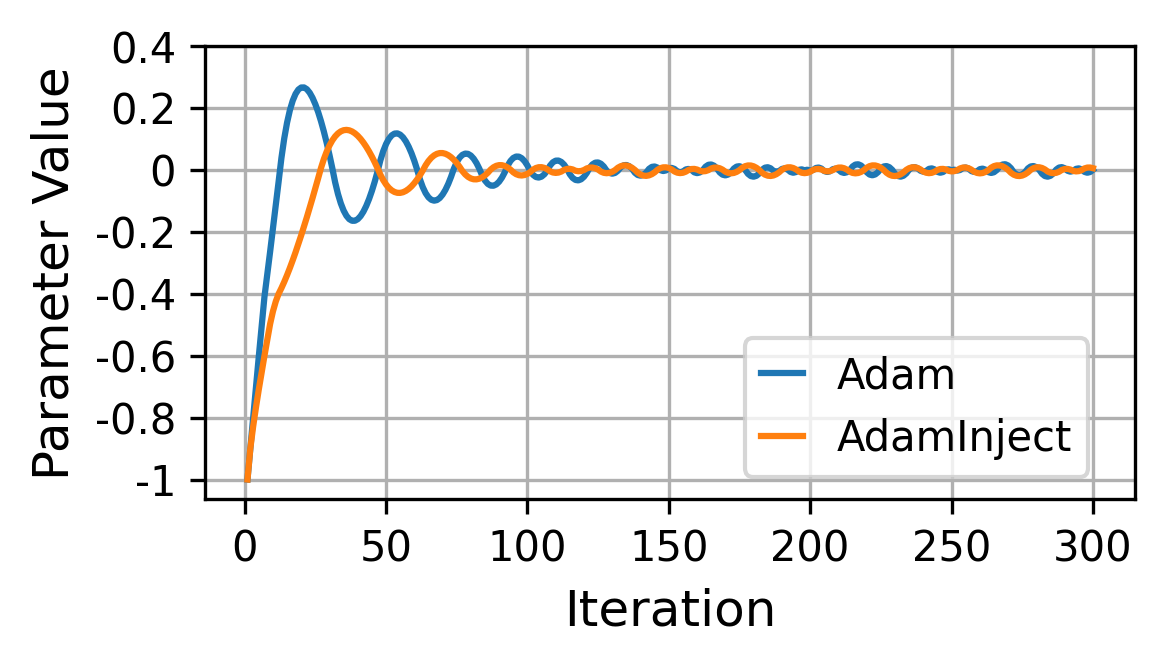}
    \includegraphics[width=0.325\textwidth]{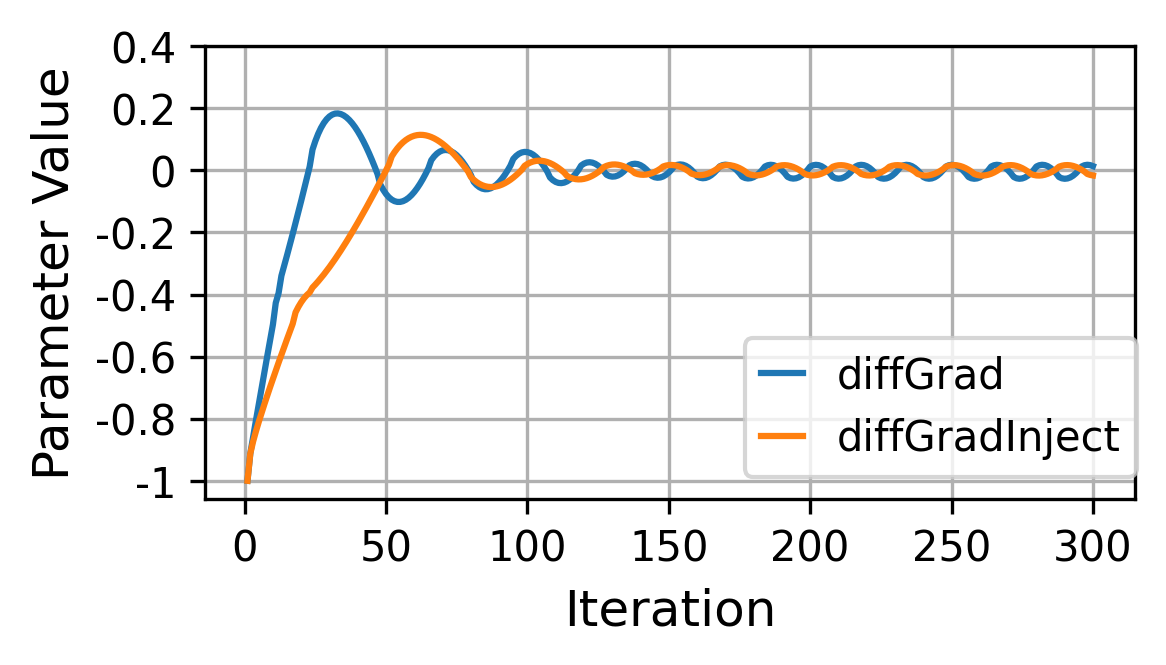}
    \caption{The empirical results computed over three synthetic, non-convex functions as toy examples. Each row corresponds to a function. The $1^{st}$ column shows the used functions. The $2^{nd}$ and $3^{rd}$ columns show the parameter value updates for $300$ iterations using Adam and diffGrad optimizers, respectively, with and without the proposed second order injection. The regression loss based objective function is used to update the parameters. The initialization is done at $x = -1$.}
    \label{fig:analytical}
\end{figure*}

\textbf{S2:} The ``large gradient, small curvature" is another scenario in optimization landscape. The gradient $|g_t|$ is higher in such regions. An ideal optimizer is expected to take the large parameter updates in such regions. The EMA of gradient (i.e., $m_t$) as well as squared gradient (i.e., $v_t$) are large. Moreover, the EMA of the proposed injected gradient (i.e., $s_t$) is also sufficiently large as $|\Delta\theta|$ is small. Hence, the SGD takes large step in this scenario. However, both the Adam and AdaInject take relatively smaller step due to large value of $\sqrt{v_t}$ in the denominator. But, we show experimently that this problem can be reduced by considering AdaBelief concept \cite{adabelief} with the proposed injection idea (i.e., AdaBeliefInject).

\textbf{S3:} The third scenario is parameter updates near ``steep and narrow” valley (i.e., minimum). It is expected for an ideal optimizer to decrease the step size for parameter updates in this scenario to avoid the overshooting as well as to reduce the oscillation near the valley. The proposed AdaInject optimizer is very beneficial in this scenario too.
The gradient $|g_t|$ is large in this scenario, hence $m_t$ and $\sqrt{v_t}$ are also large. The SGD suffers due to large value of $m_t$. This problem is reduced to a certain extent in Adam due to large value of $\sqrt{v_t}$ in the denominator. In this scenario, $|\Delta\theta|$ is large, $\Delta \theta < 0$ when $g_t > 0$ and $\Delta \theta > 0$ when $g_t < 0$, leading to $|s_t| < |m_t|$ (Note that $t$ is expected not to be the initial iterations near minimum, rather sufficiently large). Hence, the proposed AdaInject method reduces $s_t$ while enjoying the benefits of Adam (i.e., large $\sqrt{v_t}$ in denominator) leading to a reduced step size, which avoids the overshooting and oscillation near minimum to a greater extent. 
In order to show this effect using toy examples, we conduct an empirical study with the help of synthetic, non-convex functions in the next subsection.

\subsection{Empirical Analysis using Toy Examples}
We perform the empirical analysis using three synthetic, one-dimensional, non-convex functions by following the protocol of diffGrad \cite{diffgrad}. These functions are given as:
\begin{equation}
F1(x)=
\begin{cases}
(x+0.3)^2, & \text{for } x \leq 0 \\ 
(x-0.2)^2+0.05, & \text{for } x > 0 \\
\end{cases}
\end{equation}
\begin{equation}
F2(x)=
\begin{cases}
-40x-35.15, & \text{for } x \leq -0.9 \\ 
x^3+x\sin(8x)+0.85, & \text{for } x > -0.9 \\
\end{cases}
\end{equation}
\begin{equation}
F3(x)=
\begin{cases}
x^2, & \text{for } x \leq -0.5 \\
0.75+x, & \text{for } -0.5 < x \leq -0.4\\
-7x/8, & \text{for } -0.4 < x \leq 0\\
7x/8, & \text{for } 0 < x \leq 0.4\\
0.75-x, & \text{for } 0.4 < x \leq 0.5\\
x^2, & \text{for } 0.5 < x \\
\end{cases}
\end{equation}
where $-\infty < x < +\infty$ is the input. 
Functions $F1$, $F2$, and $F3$ are illustrated in Fig. \ref{fig:analytical} in the $1^{st}$ column and in the $1^{st}$, $2^{nd}$, and $3^{rd}$ rows, respectively, for $-1 < x < +1$. The parameter $x$ is initialized at $-1$. The experiment is performed by computing the regression loss as the objective function.
The $2^{nd}$ column shows the parameter values at different iterations using Adam and AdamInject optimizers. Similarly, the $3^{rd}$ column illustrates the parameter values at different iterations using diffGrad and diffGradInject optimizers. It can be noticed that Adam overshoots the minimum for both $F1$ and $F2$ functions, whereas AdamInject is able to avoid the overshooting due to the small step size caused by the proposed parameter change weighted second order moment injection in parameter update. In other cases, including Adam and AdamInject for $F3$ function and diffGrad and diffGradInject for all three functions, the effect of the proposed optimizer can be easily observed in terms of the smooth parameter updates and less oscillations near minimum by accumulating the injected momentum in an accurate direction. It is noticed that AdaInject is more effective with Adam than diffGrad as diffGrad utilizes the short-term gradient change as friction coefficient.
These results confirm that the proposed parameter change guided second moment injection leads to accurate and precise parameter updates, especially near ``steep and narrow” valley.

\begin{figure}[!t]
    \centering
    \includegraphics[width=0.241\textwidth, trim={0.7cm 0.7cm 0.2cm 0.7cm},clip] {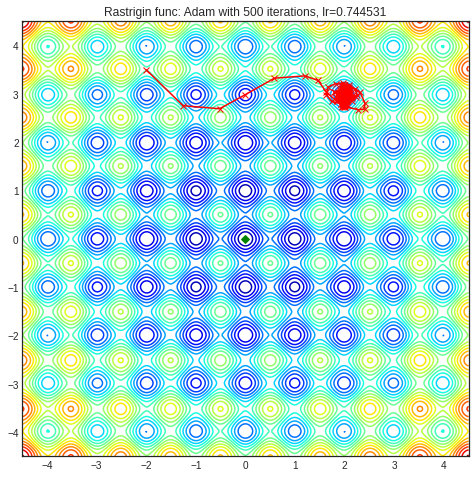}
    \includegraphics[width=0.241\textwidth, trim={0.7cm 0.7cm 0.2cm 0.7cm},clip] {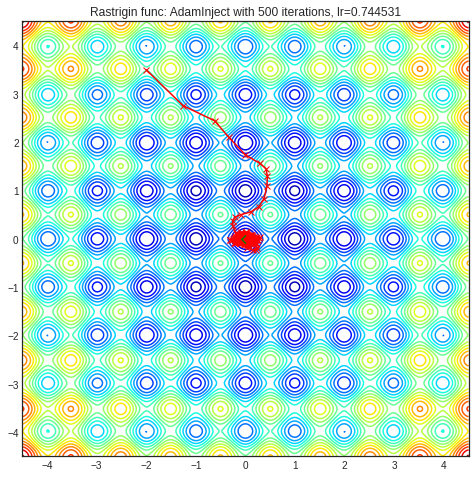}
    \includegraphics[width=0.241\textwidth, trim={0.97cm 0.7cm 0.45cm 0.7cm},clip] {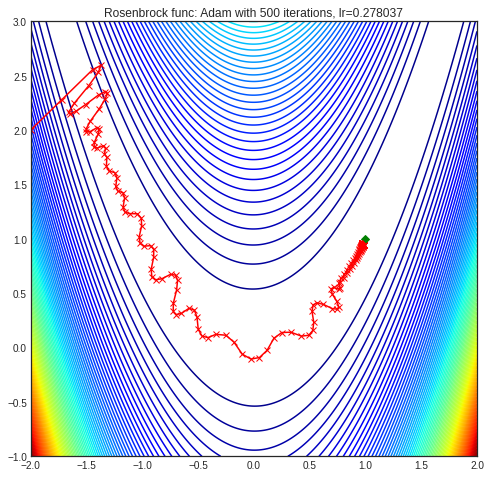}
    \includegraphics[width=0.241\textwidth, trim={0.97cm 0.7cm 0.45cm 0.7cm},clip] {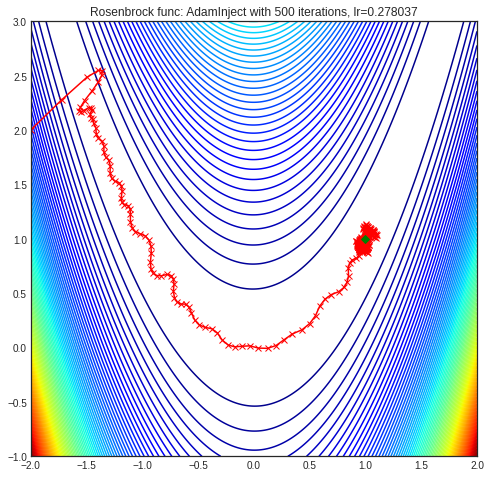}
    \caption{The optimization illustration for Rastrigin (upper row) and Rosenbrock (lower row) functions using Adam (left column) and AdamInject (right column).}
    \label{fig:rastrigin_rosenbrock}
\end{figure}

In order to demonstrate the effect of the proposed optimizer on 2-dimensional optimization, we consider non-convex Rastrigin and Rosenbrock functions\footnote{https://github.com/jettify/pytorch-optimizer}, which are widely used to show the optimization characteristics. The Rastrigin function has one global minimum at (0.0, 0.0). However, the Rosenbrock has one global minimum at (1.0. 1.0). The optimization trajectories using Adam and AdamInject optimizers under the same experimental setup are depicted in Fig. \ref{fig:rastrigin_rosenbrock}. It can be noticed that Adam is not able to converge over the Rastrigin function due to the presence of several local minima. Whereas, the AdamInject is able to converge over the Rastrigin function due to the improved parameter updates caused by the second order moment injection. It is also observed that the Adam optimizer takes more steps to reach the minimum over the Rosenbrock function due to irregular parameter updates caused by long, narrow, parabolic shaped flat valley. However, the AdamInject optimizer is able to tackle this issue and reaches the minimum in less number of steps over the Rosenbrock function.

\begin{table*}[!t]
    \caption{The experimental results of different CNNs in terms of top-1 classification error (\%) over the CIFAR10 dataset using different optimizers, without and with the proposed AdaInject. The results with the proposed approach are highlighted in bold. The improvement in the error due to the proposed injection concept is also mentioned. The highest increase for an optimizer is also highlighted in bold. The symbols $\uparrow$ and $\downarrow$ represent the improvement and degradation in \%, respectively, in the top-1 error. We follow the same convention in the results reported in Table \ref{tab:results_cifar100} and \ref{tab:results_FMNIST} also. These results are computed as the average over three independent trials.}
    \centering
    \resizebox{\textwidth}{!}{%
    \begin{tabular}{p{0.085\textwidth}|p{0.06\textwidth}|p{0.097\textwidth}|p{0.06\textwidth}|p{0.09\textwidth}|p{0.06\textwidth}|p{0.09\textwidth}|p{0.06\textwidth}|p{0.09\textwidth}}
          \hline
         CNN & \multicolumn{8}{c}{Classification error (\%) using different optimizers without and with AdaInject} \\
         \cline{2-9}
        Models & \multicolumn{2}{c|}{Adam} & \multicolumn{2}{c|}{diffGrad} & \multicolumn{2}{c|}{Radam} & \multicolumn{2}{c}{AdaBelief}\\
         \cline{2-9}
         & Adam & AdamInject & diffGrad & diffGradInject & Radam & RadamInject & AdaBelief & AdaBeliefInject\\
         \hline
        VGG16	&	7.45	&	\textbf{7.20}	($\uparrow$	3.36) &	7.24	&	\textbf{7.04}	($\uparrow$	2.76) &	7.06	&	\textbf{6.88}	($\uparrow$	2.55) &	7.29	&	\textbf{7.07}	($\uparrow$	3.02) \\
        ResNet18	&	6.46	&	\textbf{6.20}	($\uparrow$	4.02) &	6.51	&	\textbf{6.10}	($\uparrow$	6.30) &	6.18	&	\textbf{5.87}	($\uparrow$	5.02) &	6.37	&	\textbf{6.30}	($\uparrow$	1.10) \\
        SENet18	&	6.61	&	\textbf{6.29}	($\uparrow$	4.84) &	6.44	&	\textbf{6.21}	($\uparrow$	3.57) &	6.05	&	\textbf{5.83}	($\uparrow$	3.64) &	6.59	&	\textbf{6.23}	($\uparrow$	5.46) \\
        ResNet50	&	6.17	&	\textbf{5.89}	($\uparrow$	4.54) &	6.19	&	\textbf{5.73}	($\uparrow$	7.43) &	5.86	&	\textbf{5.29}	($\uparrow$	\textbf{9.73}) &	5.90	&	\textbf{5.78}	($\uparrow$	2.03) \\
        ResNet101	&	6.90	&	\textbf{6.01}	($\uparrow$	\textbf{12.90}) &	6.45	&	\textbf{5.69}	($\uparrow$	11.78) &	6.29	&	\textbf{5.76}	($\uparrow$	8.43) &	6.37	&	\textbf{6.03}	($\uparrow$	5.34) \\			
        ResNeXt29	&	6.79	&	\textbf{6.16}	($\uparrow$	9.28) &	6.83	&	\textbf{5.70}	($\uparrow$	\textbf{16.54}) &	6.00	&	\textbf{5.67}	($\uparrow$	5.50) &	6.43	&	\textbf{5.99}	($\uparrow$	\textbf{6.84}) \\			
        DenseNet121	&	6.30	&	\textbf{5.63}	($\uparrow$	10.63) &	5.90	&	\textbf{5.43}	($\uparrow$	7.97) &	5.25	&	\textbf{5.10}	($\uparrow$	2.86) &	6.05	&	\textbf{5.64}	($\uparrow$	6.78) \\
         \hline
    \end{tabular}}
\label{tab:results_cifar10}
\end{table*}

\section{Convergence Analysis}
We use the online learning framework to show the convergence property of the proposed injection based AdamInject optimizer, similar to Adam \cite{Adam}. Let's represent the unknown sequence of convex cost functions as $f_1(\theta)$, $f_2(\theta)$,$...$, $f_T(\theta)$. We want to estimate parameter $\theta_t$ at each iteration $t$ and assess over $f_t(\theta)$. The regret bound is commonly used in such scenarios to assess the algorithm where the information of the sequence is not known in advance. The sum of the difference between all the previous online guesses $f_t(\theta_t)$ and the best fixed point parameter $f_t(\theta^*)$ from a feasible set $\chi$ of all the previous iterations are used to compute the regret bound. The regret bound is given as,
\begin{equation}
R(T)=\sum_{t=1}^{T}{[f_t(\theta_t)-f_t(\theta^*)]}
\end{equation}
where $\theta^*=\mbox{arg }\mbox{min}_{\theta \in \chi }\sum_{t=1}^{T}{f_t(\theta)}$. The regret bound for the proposed injection based AdamInject is noticed as $O(\sqrt{T})$ which is the same as Adam and is comparable to general convex online learning approaches. We provide the proof in the Supplementary. We consider $g_{t,i}$ as the gradient for the $i^{th}$ element in the $t^{th}$ iteration, $g_{1:t,i}=[g_{1,i},g_{2,i},...,g_{t,i}] \in \mathbb{R}^t$ is the gradient vector in the $i^{th}$ dimension up to $t^{th}$ iterations, and $\gamma \triangleq \frac{\beta_1^2}{\sqrt{\beta_2}}$. 

\begin{theorem}
\textit{Assume that the gradients for function $f_t$ (i.e., $||g_{t,\theta}||_2 \leq G$ and $||g_{t,\theta}||_{\infty} \leq G_{\infty}$) are bounded for all $\theta \in R^d$. Let also consider that the bounded distance is generated by the proposed optimizer between any $\theta_t$ (i.e., $||\theta_n-\theta_m||_2 \leq D$ and $||\theta_n-\theta_m||_\infty \leq D_\infty$ for any $m,n\in\{1,...,T\}$). Let $\gamma \triangleq \frac{\beta_1^2}{\sqrt{\beta_2}}$, $\beta_1,\beta_2 \in [0,1)$ satisfy $\frac{\beta_1^2}{\sqrt{\beta_2}} < 1$, $\alpha_t=\frac{\alpha}{\sqrt{t}}$, and $\beta_{1,t}=\beta_1\lambda^{t-1},\lambda \in (0,1)$ with $\lambda$ is around $1$, e.g $1-10^{-8}$. For all $T \geq 1$, the proposed injection based AdamInject shows the following guarantee as derived in the Supplemetary:} 
\begin{dmath}
R(T) \leq \frac{D^2}{\alpha(1-\beta_1)}\sum_{i=1}^{d}{\sqrt{T\hat{v}_{T,i}}} 
+ \frac{2\alpha(1+\beta_1) G_\infty}{(1-\beta_1)\sqrt{1-\beta_2}(1-\gamma)^2}\sum_{i=1}^{d}{||g_{1:T,i}||_2} 
+ \sum_{i=1}^{d}{\frac{D_{\infty}^{2}G_{\infty}\sqrt{1-\beta_2}}{\alpha (1-\beta_1)(1-\lambda)^2}}
+ 4D_\infty G_\infty^2 \sum_{i=1}^{d}{||g_{1:T,i}||_2^2}
\end{dmath}
\end{theorem}
The aggregation terms over the dimension ($d$) can be very small as compared to the corresponding upper bounds, such as $\sum_{i=1}^{d}{||g_{1:T,i}||_2}<< dG_\infty\sqrt{T}$, $\sum_{i=1}^{d}{||g_{1:T,i}||_2^2}<< dG_\infty\sqrt{T}$ and $\sum_{i=1}^{d}{\sqrt{T\hat{v}_{T,i}}} << dG_\infty\sqrt{T}$. The adaptive methods such as the proposed optimizers and Adam show the upper bound as $O(\log d\sqrt{T})$, which is better than $O(\sqrt{dT})$ of non-adaptive optimizers. By following the convergence analysis of Adam \cite{Adam}, we also use the decay of $\beta_{1,t}$.

We show the convergence of average regret of AdamInject in below corollary with the help of the above theorem and $\sum_{i=1}^{d}{||g_{1:T,i}||_2}<< dG_\infty\sqrt{T}$, $\sum_{i=1}^{d}{||g_{1:T,i}||_2^2}<< dG_\infty\sqrt{T}$ and $\sum_{i=1}^{d}{\sqrt{T\hat{v}_{T,i}}} << dG_\infty\sqrt{T}$.
\begin{corollary}
\textit{Assume that the gradients for function $f_t$ (i.e., $||g_{t,\theta}||_2 \leq G$ and $||g_{t,\theta}||_{\infty} \leq G_{\infty}$) are bounded for all $\theta \in R^d$. Let also consider that the bounded distance is generated by the proposed optimizer between any $\theta_t$ (i.e., $||\theta_n-\theta_m||_2 \leq D$ and $||\theta_n-\theta_m||_\infty \leq D_\infty$ for any $m,n\in\{1,...,T\}$). For all $T \geq 1$, the proposed injection based AdamInject optimizer shows the following guarantee:} 
\begin{equation}
\frac{R(T)}{T}=O(\frac{1}{\sqrt{T}}). 
\end{equation}
Thus, $\lim_{T\rightarrow\infty}\frac{R(T)}{T}=0$.
\end{corollary}

\begin{table*}[!t]
    \caption{The experimental results of different CNNs in terms of top-1 classification error (\%) over the CIFAR100 dataset using different optimizers, without and with the proposed AdaInject. These results are computed as the average over three independent trials.}
    \centering
     \resizebox{\textwidth}{!}{%
     \begin{tabular}{p{0.085\textwidth}|p{0.06\textwidth}|p{0.097\textwidth}|p{0.06\textwidth}|p{0.09\textwidth}|p{0.06\textwidth}|p{0.09\textwidth}|p{0.06\textwidth}|p{0.09\textwidth}}
        \hline
         CNN & \multicolumn{8}{c}{Classification error (\%) using different optimizers without and with AdaInject} \\
         \cline{2-9}
        Models & Adam & AdamInject & diffGrad & diffGradInject & Radam & RadamInject & AdaBelief & AdaBeliefInject\\
         \hline
         VGG16	&	32.71	&	\textbf{31.81}	($\uparrow$	2.75) &	31.81	&	\textbf{30.80}	($\uparrow$	3.18) &	\textbf{29.31}	&	30.07	($\downarrow$	2.59) &	31.08	&	\textbf{30.04}	($\uparrow$	3.35) \\			
        ResNet18	&	28.91	&	\textbf{27.28}	($\uparrow$	5.64) &	26.50	&	\textbf{26.23}	($\uparrow$	1.02) &	26.78	&	\textbf{25.84}	($\uparrow$	3.51) &	27.28	&	\textbf{26.31}	($\uparrow$	3.56) \\
        SENet18	&	29.15	&	\textbf{28.74}	($\uparrow$	1.41) &	28.60	&	\textbf{27.64}	($\uparrow$	3.36) &	27.66	&	\textbf{26.63}	($\uparrow$	3.72) &	26.90	&	\textbf{26.52}	($\uparrow$	1.41) \\
        ResNet50	&	28.12	&	\textbf{25.44}	($\uparrow$	9.53) &	24.94	&	\textbf{24.18}	($\uparrow$	3.05) &	25.05	&	\textbf{24.13}	($\uparrow$	3.67) &	24.47	&	\textbf{24.25}	($\uparrow$	0.90) \\
        ResNet101	&	25.78	&	\textbf{23.98}	($\uparrow$	6.98) &	26.58	&	\textbf{24.17}	($\uparrow$	\textbf{9.07}) &	25.74	&	\textbf{23.83}	($\uparrow$	7.42) &	\textbf{24.12}	&	24.24	($\downarrow$	0.50) \\
        ResNeXt29	&	28.78	&	\textbf{24.96}	($\uparrow$	\textbf{13.27}) &	25.47	&	\textbf{24.53}	($\uparrow$	3.69) &	24.66	&	\textbf{22.74}	($\uparrow$	7.79) &	24.61	&	\textbf{23.63}	($\uparrow$	\textbf{3.98}) \\			
        DenseNet121	&	26.40	&	\textbf{24.33}	($\uparrow$	7.84) &	24.14	&	\textbf{23.66}	($\uparrow$	1.99) &	25.17	&	\textbf{23.06}	($\uparrow$	\textbf{8.38}) &	24.68	&	\textbf{24.06}	($\uparrow$	2.51) \\
         \hline
    \end{tabular}}
\label{tab:results_cifar100}
\end{table*}

\begin{table*}[!t]
\caption{The experimental results of different CNNs in terms of top-1 classification error (\%) over the FashionMNIST dataset using different optimizers, without and with the proposed AdaInject. These results are computed as the average over three independent trials.}
    \centering
    \resizebox{\textwidth}{!}{%
    \begin{tabular}{p{0.085\textwidth}|p{0.06\textwidth}|p{0.097\textwidth}|p{0.06\textwidth}|p{0.09\textwidth}|p{0.06\textwidth}|p{0.09\textwidth}|p{0.06\textwidth}|p{0.09\textwidth}}
    \hline
         CNN & \multicolumn{8}{c}{Classification error (\%) using different optimizers without and with AdaInject} \\
         \cline{2-9}
        Models & Adam & AdamInject & diffGrad & diffGradInject & Radam & RadamInject & AdaBelief & AdaBeliefInject\\
         \hline
        VGG16	&	5.15	&	\textbf{5.01}	($\uparrow$	2.72) &	5.13	&	\textbf{5.03}	($\uparrow$	1.95) &	5.11	&	\textbf{5.07}	($\uparrow$	0.78) &	5.12	&	\textbf{4.97}	($\uparrow$	2.93) \\			
        ResNet18	&	4.76	&	\textbf{4.74}	($\uparrow$	0.42) &	4.82	&	\textbf{4.65}	($\uparrow$	3.53) &	4.78	&	\textbf{4.67}	($\uparrow$	2.30) &	4.95	&	\textbf{4.75}	($\uparrow$	4.04) \\			
        SENet18	&	5.14	&	\textbf{4.95}	($\uparrow$	3.70) &	5.11	&	\textbf{4.79}	($\uparrow$	6.26) &	5.08	&	\textbf{4.79}	($\uparrow$	5.71) &	5.06	&	\textbf{4.91}	($\uparrow$	2.96) \\			
        ResNet50	&	5.10	&	\textbf{4.76}	($\uparrow$	6.67) &	4.93	&	\textbf{4.77}	($\uparrow$	3.25) &	4.98	&	\textbf{4.84}	($\uparrow$	2.81) &	5.10	&	\textbf{4.78}	($\uparrow$	6.27) \\			
        ResNet101	&	4.94	&	\textbf{4.65}	($\uparrow$	5.87) &	5.05	&	\textbf{4.73}	($\uparrow$	6.34) &	4.91	&	\textbf{4.64}	($\uparrow$	5.50) &	5.21	&	\textbf{4.69}	($\uparrow$	\textbf{9.98}) \\			
        ResNeXt29	&	6.16	&	\textbf{5.59}	($\uparrow$	\textbf{9.25}) &	5.92	&	\textbf{5.16}	($\uparrow$	\textbf{12.84}) &	5.78	&	\textbf{5.37}	($\uparrow$	\textbf{7.09}) &	5.25	&	\textbf{4.90}	($\uparrow$	6.67) \\
        DenseNet121	&	4.88	&	\textbf{4.69}	($\uparrow$	3.89) &	4.77	&	\textbf{4.70}	($\uparrow$	1.47) &	4.89	&	\textbf{4.68}	($\uparrow$	4.29) &	4.68	&	\textbf{4.56}	($\uparrow$	2.56) \\			\hline
    \end{tabular}}
\label{tab:results_FMNIST}
\end{table*}

\section{Experimental Analysis}
In this section, first we describe the experimental setup. Then, we present the detailed results using different optimizers. Finally, we analyze effects of the hyperparameters.

\subsection{Experimental Setup}
We use a wide range of CNN models (i.e., VGG16 \cite{VggNet}, ResNet18, ResNet50, ResNet101 \cite{ResNet}, SENet18 \cite{senet}, ResNeXt29 \cite{resnext} and DenseNet121 \cite{densenet}) to test the suitability of the proposed AdaInject concept for optimizers.
We follow the publicly available Pytorch implementation\footnote{https://github.com/kuangliu/pytorch-cifar} of these CNN models. For ResNeXt29 model, we set the cardinality as $4$ and bottleneck width as $64$.
We train all the CNN models using all the optimizers under the same experimental setup. The training is performed for $100$ epochs with a batch size of $64$ for CIFAR10/100 and FashionMNIST and $256$ for TinyImageNet dataset. The learning rate is set to $0.001$ for the first $80$ epochs and $0.0001$ for the last $20$ epochs. Different computers are used for the experiments, including Google colaboratory\footnote{https://colab.research.google.com/}. We performed a random crop and random horizontal flip over training data. The normalization is performed for both training and test data.

In order to demonstrate the efficacy of the proposed AdaInject based optimizers experimentally, we use four benchmark object recognition dataset, including CIFAR10 \cite{cifar}, CIFAR100 \cite{cifar}, FashionMNIST \cite{fmnist}, and TinyImageNet\footnote{http://cs231n.stanford.edu/tiny-imagenet-200.zip} \cite{tinyimagenet}. We use CIFAR and FashionMNIST datasets directly from the PyTorch library. 
CIFAR10 dataset consists of a total $60,000$ images of dimension $32 \times 32 \times 3$ from $10$ object classes with $6,000$ images per class. The training set contains $50,000$ images with $5,000$ images per class and the test set contains $10,000$ images with $1,000$ images per class in CIFAR10. 
CIFAR100 dataset contains all the images of CIFAR10, but is categorized into $100$ classes. Thus, CIFAR100 dataset contains $50,000$ training images and $10,000$ test images with $500$ and $100$ images per class, respectively. 
FashionMNIST dataset contains $70,000$ labeled fashion images of dimension $28 \times 28$ from $10$ categories. The training and test sets consist of $60,000$ and $10,000$ images, respectively. TinyImageNet dataset \cite{tinyimagenet} is a subset of the large-scale visual recognition ImageNet challenge \cite{imagenet}. This dataset consists of the images from $200$ object classes with $1,00,000$ images in the training set (i.e., $500$ images in each class) and $10,000$ images in the validation set  (i.e., $50$ images in each class).

\begin{table*}[!t]
\caption{The experimental results of VGG16, ResNet18, and SENet models in terms of top-1 classification accuracy (\%) over the TinyImageNet dataset using different optimizers. These results are computed as the average over three independent trials.}
    \centering
    \resizebox{\textwidth}{!}{%
    \begin{tabular}{p{0.07\textwidth}|p{0.06\textwidth}|p{0.09\textwidth}|p{0.06\textwidth}|p{0.09\textwidth}|p{0.06\textwidth}|p{0.09\textwidth}|p{0.06\textwidth}|p{0.09\textwidth}}          
    \hline
         CNN & \multicolumn{8}{c}{Accuracy (\%) using different optimizers without and with AdaInject} \\
         \cline{2-9}
        Models & Adam & AdamInject & diffGrad & diffGradInject & Radam & RadamInject & AdaBelief & AdaBeliefInject\\
         \hline
        VGG16	&	44.05	&	\textbf{44.58}	($\uparrow$	1.20) &	46.00	&	\textbf{47.18}	($\uparrow$	2.57) &	45.92	&	\textbf{46.38}	($\uparrow$	1.00) &	47.88	&	\textbf{48.25}	($\uparrow$	0.77) \\
        ResNet18	&	50.58	&	\textbf{51.90}	($\uparrow$	2.61) &	52.04	&	\textbf{52.37}	($\uparrow$	0.63) &	52.12	&	\textbf{52.50}	($\uparrow$	0.73) &	52.05	&	\textbf{52.74}	($\uparrow$	1.33) \\
        SENet18	&	48.04	&	\textbf{49.52}	($\uparrow$	3.08) &	49.51	&	\textbf{50.28}	($\uparrow$	1.56) &	50.73	&	\textbf{51.01}	($\uparrow$	0.55) &	51.76	&	\textbf{51.94}	($\uparrow$	0.35) \\
        \hline
    \end{tabular}}
\label{tab:results_tinyimagenet}
\end{table*}

\subsection{Experimental Results}
We compare the performance using four recent state-of-the-art adaptive gradient descent optimizers (i.e., Adam \cite{Adam}, diffGrad \cite{diffgrad}, Radam \cite{radam} and AdaBelief \cite{adabelief}), without and with the proposed injection approach. We consider VGG16 \cite{VggNet}, ResNet18, ResNet50, ResNet101 \cite{ResNet}, SENet18 \cite{senet}, ResNeXt29 \cite{resnext} and DenseNet121 \cite{densenet} CNN models.
The experimental results over the CIFAR10 dataset are depicted in Table \ref{tab:results_cifar10} in terms of the error rate. It is observed that the performance of all CNN models is improved with AdaInject based optimizers as compared to its performance with corresponding vanilla optimizers. The RadamInject optimizer leads to best performance using the DenseNet121 model with a $5.10\%$ error rate in classification. The highest improvement is reported by the ResNeXt29 model using diffGradInject. Moreover, the performance of the ResNeXt29 model is also significantly improved using AdaBeliefInject. In general, we observe better performance gain by heavy CNN models. 

The results over the CIFAR100 dataset are illustrated in Table \ref{tab:results_cifar100}. The best performance of $77.26\%$ accuracy is achieved by the RadamInject optimizer using the ResNeXt29 model. The performance of ResNeXt29 is improved significantly using the proposed injection for optimizers with highest improvement by AdamInject. The results due to the proposed injection based optimizers are improved using all the CNN models except RadamInject using VGG16 and AdaBeliefInject using ResNet101. Note that Radam does not use second order moment if rectification criteria is not met and AdaBelief reduces the second order moment. These could be the possible reasons that the performance of RadamInject and AdaBeliefInject is marginally down in some cases.
A very similar trend is also noticed over FashionMNIST (FMNIST) dataset in Table \ref{tab:results_FMNIST}, where the performance using the proposed approach is improved in all the cases. The best accuracy of $95.44\%$ is observed for the AdaBeliefInject optimizer using the DenseNet121 model. An outstanding improvement in top-1 error is perceived for the ResNeXt29 model over the FashionMNIST dataset using the optimizers with the proposed AdaInject concept. The performance of other models is also significantly improved due to the proposed injection approach.

\begin{table}[!t]
    \caption{Accuracy (\%) using AdamInject optimizer with different values of $k$. Results are computed as the average over three independent trials. Note that best and second best results are highlighted in Bold and Underline, respectively.}
    \centering
    \resizebox{\columnwidth}{!}{%
    \begin{tabular}{p{0.041\textwidth}p{0.069\textwidth}p{0.021\textwidth}p{0.02\textwidth}p{0.02\textwidth}p{0.02\textwidth}p{0.02\textwidth}p{0.02\textwidth}p{0.02\textwidth}p{0.022\textwidth}}
    \hline
    Model & Dataset & \textit{k}=1 & \textit{k}=2 & \textit{k}=3 & \textit{k}=4 & \textit{k}=5 & \textit{k}=10 & \textit{k}=20 & \textit{k}=50 \\\hline
    \multirow{4}{*}{VGG16} & CIFAR10 & 92.68 & \textbf{92.80} & \underline{92.78} & 92.50 & 92.61 & 91.85 & 90.75 & 87.76\\
    & CIFAR100 & 67.04 & 68.19 & 67.53 & \underline{68.29} & \textbf{68.48} & 68.25 & 66.97 & 61.08\\
    & MNIST & \underline{94.94} & \textbf{94.99} & 94.91 & 94.91 & 94.93 & 94.8 & 94.47 & 94.15\\
    & TinyImageNet & 42.46 & \textbf{44.58} & 43.14 & 42.71 & \underline{44.21} & 44.04 & 44.05 & 40.55\\
    \hline
    \multirow{4}{*}{ResNet18} & CIFAR10 & 93.71 & \underline{93.80} & \textbf{93.88} & 93.75 & 93.76 & 93.41 & 92.24 & 89.36\\
    & CIFAR100 & 71.97 & 72.72 & \underline{73.30} & 73.26 & \textbf{73.59} & 72.76 & 69.99 & 64.43\\
    & MNIST & 95.15 & \underline{95.26} & 95.19 & \textbf{95.33} & 95.18 & 95.16 & 94.84 & 94.52\\
    & TinyImageNet & 49.09 & \textbf{51.90} & 48.47 & 50.43 & \underline{51.17} & 50.11 & 49.37 & 43.64\\
    \hline
    \end{tabular}}
\label{tab:results_k_adaminject}
\end{table}

We also perform the experiment over the TinyImageNet dataset using VGG16, ResNet18 and SENet18 models and show the results in terms of the classification accuracy in \% in Table \ref{tab:results_tinyimagenet} for different optimizers with and without the proposed injection concept. It is observed from this experiment that the proposed approach is able to improve the performance of the existing optimizers over large scale dataset as well.
These results confirm that the proposed injection updates the parameter in an optimal way by utilizing the short-term parameter update information with second order moment.

\subsection{Effect of Injection Hyperparameter ($k$)}
In the previous results, we use the value of the injection hyperparameter ($k$) as $2$. We show a performance comparison by considering the value of $k$ as $1, 2, 3, 4, 5$, $10$, $20$, and $50$ in Table \ref{tab:results_k_adaminject}. The results are presented using the AdamInject optimizer for VGG16 and ResNet18 models over the CIFAR10, CIFAR100, FMNIST, and TinyImageNet datasets. 
It is noticed that $k=2$ is better suitable for the VGG16 model on CIFAR10, MNIST and TinyImageNet datasets. Moreover, the accuracy using $k=2$ is also either best or second best for the ResNet18 model on CIFAR10, MNIST and TinyImageNet datasets. It is also evident that the results on fine-grained CIFAR100 dataset are best using $k=5$ for both VGG16 and ResNet18 models. It is suggested to consider the value of $k \in \{2,3,4,5\}$. The original selection of the value of $k$ as 2 is also justified from this analysis.

\begin{table}[!t]
    \caption{Accuracy (\%) using AdamInject optimizer with different batch size (BS) and learning rate (LR). Results are computed as the average over three independent trials. Note that best results are highlighted in Bold.}
    \centering
    \begin{tabular}{p{0.05\textwidth}|p{0.08\textwidth}|p{0.028\textwidth}p{0.028\textwidth}p{0.028\textwidth}|p{0.028\textwidth}p{0.028\textwidth}p{0.028\textwidth}}
    \hline
    \multirow{2}{*}{Model} & \multirow{2}{*}{Dataset} & \multicolumn{3}{c|}{Batch Size (BS)} & \multicolumn{3}{c}{Learning Rate (LR)} \\\cline{3-5}\cline{6-8}
    & & 32 & 64 & 128 & 0.0001 & 0.001 & 0.01 \\\hline
    \multirow{4}{*}{VGG16} & CIFAR10 & 92.46 & \textbf{92.80} & 92.45 & 91.16 & \textbf{92.80} & 92.45\\
    & CIFAR100 & 67.00 & \textbf{68.19} & 68.03 & 66.92 & \textbf{68.19} & 66.97\\
    & MNIST & 94.90 & \textbf{94.99} & 94.96 & 94.75 & \textbf{94.99} & 94.85\\
    & TinyImageNet & 41.67 & \textbf{44.58} & 42.69 & \textbf{45.18} & 44.58 & 39.33\\
    \hline
    \multirow{4}{*}{ResNet18} & CIFAR10 & 93.94 & \textbf{94.13} & 93.71 & 92.36 & \textbf{94.13} & 93.65\\
    & CIFAR100 & 73.05 & \textbf{74.16} & 72.76 & 70.79 & \textbf{74.16} & 67.31\\
    & MNIST & 95.27 & \textbf{95.33} & 95.25 & 95.13 & \textbf{95.33} & 95.26\\
    & TinyImageNet & 49.63 & \textbf{51.90} & 50.18 & 49.30 & \textbf{51.90} & 45.96\\
    \hline
    \end{tabular}
\label{tab:bs_lr}
\end{table}

\subsection{Effect of Batch Size and Learning Rate}
In the previous experiments, the batch size (BS) and learning rate (LR) was set to 64 and 0.001, respectively. In this experiment, we analyze the impact of batch size and learning rate as detailed in Table \ref{tab:bs_lr}. The results are reported for VGG16 and ResNet18 models on CIFAR10, CIFAR100, MNIST and TinyImageNet datasets. The batch size is considered as 32, 64 and 128, respectively. It is evident from the results that the batch size as 64 is better suitable with the proposed AdamInject optimizer in all the cases. The learning rate is considered as 0.0001, 0.001 and 0.01, respectively. Note that the learning rate is divided by 10 once in all the cases after 80 epochs of training for a fair comparison. It is noticed that the proposed optimizer performs best for 0.001 learning rate in almost all the cases. This analysis confirms the suitability of original batch size (i.e., 64) and learning rate (i.e., 0.001) choices used for the experiments.

\section{Conclusion}
In this paper, we present a novel and generic injection based EMA of gradients for parameter update by utilizing the parameter change information along with the second order moment. 
The proposed injection approach leads to an accurate and precise update by performing smaller updates near minimum to avoid the overshooting as well as oscillation and reasonably higher updates in the small curvature regions. The effect of the proposed injection based optimizers is observed using toy examples. The convergence property of the proposed optimizer is also analyzed. The object recognition results for different CNN models over benchmark datasets using four optimizers show the superiority of the proposed injection concept. It is noticed that the injection hyperparameter as 2 yeilds to better results in majority of the cases using the AdamInject optimizer. It is also noted that the batch size as 64 and learning rate as 0.001 are better suitable with the proposed AdamInject optimizer. The intuitive explanation, empirical, convergence, and experimental analyses are evident that the proposed injection based optimizers lead to better optimization of CNNs by avoiding the overshooting of the minimum and reducing the oscillation near minimum to a greater extent.

{\small
\bibliographystyle{IEEEtran}
\bibliography{References}
}

\begin{IEEEbiography}[{\includegraphics[width=1in,height=1.25in,clip,keepaspectratio]{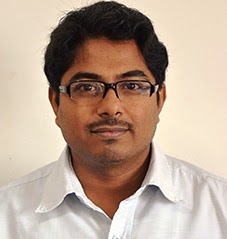}}]{Shiv Ram Dubey}
is with the Indian Institute of Information Technology (IIIT), Allahabad since July 2021, where he is currently the Assistant Professor of Information Technology. He was with IIIT Sri City as Assistant Professor from Dec 2016 to July 2021 and Research Scientist from June 2016 to Dec 2016. He received the PhD degree from IIIT Allahabad in 2016. Before that, from 2012 to 2013, he was a Project Officer at Indian Institute of Technology (IIT), Madras.
He was a recipient of several awards including Best PhD Award in PhD Symposium at IEEE-CICT2017, Early Career Research Award from SERB, Govt. of India and NVIDIA GPU Grant Award Twice from NVIDIA.
His research interest includes Computer Vision and Deep Learning.
\end{IEEEbiography}

\begin{IEEEbiography}[{\includegraphics[width=1in,height=1.25in,clip,keepaspectratio]{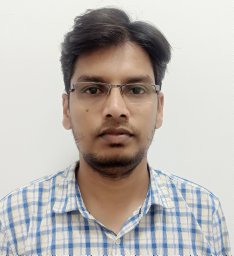}}]{S.H. Shabbeer Basha} is associated with PathPartner Technology Pvt. Ltd., Bangalore, as a lead computer vision engineer. At PathPartner, he is involved in R\&D activitis on neural network compression and deep learning. He received the PhD degree from IIIT Sri City. His research interests include Computer Vision, Deep Learning, Deep Model Compression, Unsupervised Domain Adaptation, Transfer Learning, and Multi-Task Learning.
\end{IEEEbiography}

\begin{IEEEbiography}[{\includegraphics[width=1in,height=1.25in,clip,keepaspectratio]{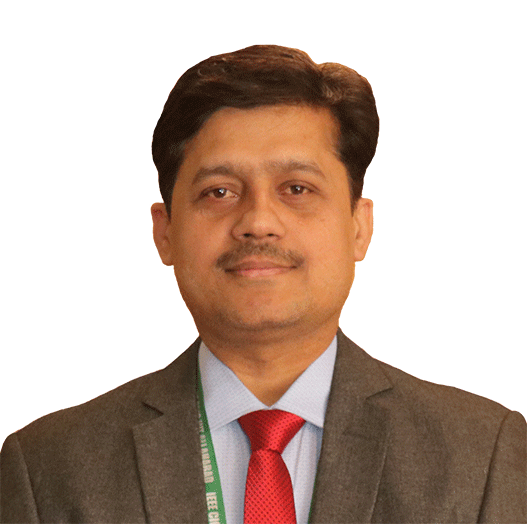}}]{Satish Kumar Singh}
is with the Indian Institute of Information Technology Allahabad, as an Associate Professor from 2013 and heading the Computer Vision and Biometrics Lab (CVBL). Earlier, he served at Jaypee University of Engineering and Technology Guna, India from 2005 to 2012. His areas of interest include Image Processing, Computer Vision, Biometrics, Deep Learning, and Pattern Recognition. 
Dr. Singh is proactively offering his volunteer services to IEEE for the last many years in various capacities. He is the senior member of IEEE. Presently Dr. Singh is Section Chair IEEE Uttar Pradesh Section (2021-2022) and a member of IEEE India Council (2021). He also served as the Vice-Chair, Operations, Outreach and Strategic Planning of IEEE India Council (2020) \& Vice-Chair IEEE Uttar Pradesh Section (2019 \& 2020). Prior to that Dr. Singh was Secretary, IEEE UP Section (2017 \& 2018), Treasurer, IEEE UP Section (2016 \& 2017), Joint Secretary, IEEE UP Section (2015), Convener Web and Newsletters Committee (2014 \& 2015). 
Dr. Singh is also the technical committee affiliate of IEEE SPS IVMSP and MMSP and presently the Chair of IEEE Signal Processing Society Chapter of Uttar Pradesh Section. 
\end{IEEEbiography}

\begin{IEEEbiography}[{\includegraphics[width=1in,height=1.25in,clip,keepaspectratio]{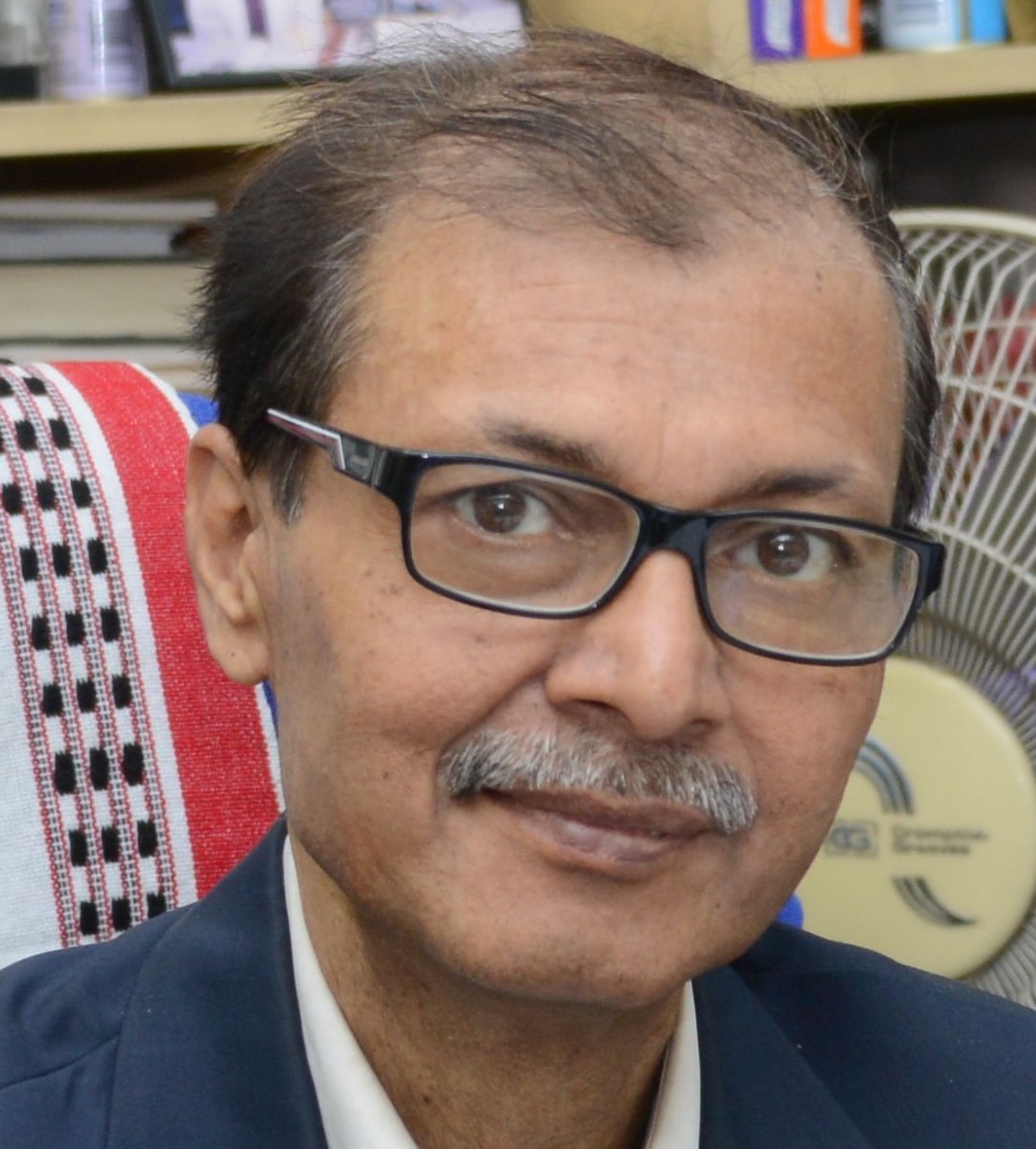}}]{Bidyut Baran Chaudhuri}
received the PhD degree from IIT Kanpur, in 1980. He was a Leverhulme Postdoctoral Fellow with Queen’s University, U.K., from 1981 to 1982. He joined the Indian Statistical Institute, in 1978, where he worked as an INAE Distinguished Professor and a J C Bose Fellow at Computer Vision and Pattern Recognition Unit. He is now affiliated to Techno India University, Kolkata as Pro-Vice Chancellor (Academic). His research interests include Pattern Recognition, Image Processing, Computer Vision, and Deep learning, etc. He pioneered the first workable OCR system for printed Indian scripts Bangla, Assamese and Devnagari. He also developed computerized \textit{Bharati Braille system} with speech synthesizer and has done statistical analysis of Indian language.  Prof. Chaudhuri received Leverhulme fellowship award, Sir J. C. Bose Memorial Award, M. N. Saha Memorial Award, Homi Bhabha Fellowship, Dr. Vikram Sarabhai Research Award, C. Achuta Menon Award, Homi Bhabha Award: Applied Sciences, Ram Lal Wadhwa Gold Medal, Jawaharlal Nehru Fellowship, J C Bose fellowship, Om Prakash Bhasin Award, etc. Prof. Chaudhuri is the fellow of INSA, NASI, INAE, IAPR, The World Academy of Sciences (TWAS) and life fellow of IEEE (2015).
\end{IEEEbiography}

\section*{Supplementary}
\subsection*{A. Convergence Proof}
\begin{theorem}
\textit{Assume that the gradients for function $f_t$ (i.e., $||g_{t,\theta}||_2 \leq G$ and $||g_{t,\theta}||_{\infty} \leq G_{\infty}$) are bounded for all $\theta \in R^d$. Let also consider that the bounded distance is generated by the proposed optimizer between any $\theta_t$ (i.e., $||\theta_n-\theta_m||_2 \leq D$ and $||\theta_n-\theta_m||_\infty \leq D_\infty$ for any $m,n\in\{1,...,T\}$). Let $\gamma \triangleq \frac{\beta_1^2}{\sqrt{\beta_2}}$, $\beta_1,\beta_2 \in [0,1)$ satisfy $\frac{\beta_1^2}{\sqrt{\beta_2}} < 1$, $\alpha_t=\frac{\alpha}{\sqrt{t}}$, and $\beta_{1,t}=\beta_1\lambda^{t-1},\lambda \in (0,1)$ with $\lambda$ is around $1$, e.g $1-10^{-8}$. For all $T \geq 1$, the proposed injection based AdamInject optimizer shows the following guarantee:} 
\begin{dmath}
R(T) \leq \frac{D^2}{\alpha(1-\beta_1)}\sum_{i=1}^{d}{\sqrt{T\hat{v}_{T,i}}} 
+ \frac{2\alpha(1+\beta_1) G_\infty}{(1-\beta_1)\sqrt{1-\beta_2}(1-\gamma)^2}\sum_{i=1}^{d}{||g_{1:T,i}||_2} 
+ \sum_{i=1}^{d}{\frac{D_{\infty}^{2}G_{\infty}\sqrt{1-\beta_2}}{\alpha (1-\beta_1)(1-\lambda)^2}}
+ 4D_\infty G_\infty^2 \sum_{i=1}^{d}{||g_{1:T,i}||_2^2}
\end{dmath}

\begin{proof}[Proof]
Following can be written from Lemma 10.2 of Adam \cite{Adam},
$$
f_t(\theta_t)-f_t(\theta^*) \leq g_t^T(\theta_t-\theta^*) = \sum_{i=1}^{d}{g_{t,i}(\theta_{t,i}-\theta_{,i}^*)}
$$
Following can be also written by utilizing the update formula of the proposed injection with Adam (i.e., AdamInject) with $k=2$ and after discarding $\epsilon$, 
\begin{dmath}
\theta_{t+1} =\theta_t-\frac{\alpha_t \hat{s}_t}{\sqrt[]{\hat{v}_{t}}}
 =\theta_t-\frac{\alpha_t}{(1-\beta_1^t)} \Big(\frac{\beta_{1,t}}{\sqrt[]{\hat{v}_{t}}}s_{t-1} + \frac{(1-\beta_{1,t})}{\sqrt[]{\hat{v}_{t}}} \frac{(g_t + \Delta \theta g_t^2)}{2}\Big)
\end{dmath}
where $\beta_{1,t}$ is the $1^{st}$ order moment coefficient at $t^{th}$ iteration. \\
We can write the following w.r.t. the $i^{th}$ dimension of parameter vector $\theta_t \in R^d$,
\begin{dmath}
(\theta_{t+1,i}-\theta_{,i}^*)^2=(\theta_{t,i}-\theta_{,i}^*)^2-\frac{2\alpha_t}{1-\beta_1^t}
\Big(\frac{\beta_{1,t}}{\sqrt[]{\hat{v}_{t,i}}}s_{t-1,i} + \frac{(1-\beta_{1,t})}{\sqrt[]{\hat{v}_{t,i}}} \frac{(g_{t,i} + \Delta \theta g_{t,i}^2)}{2} \Big)(\theta_{t,i}-\theta_{,i}^*)
+\alpha_t^2 (\frac{\hat{s}_{t,i}}{\sqrt{\hat{v}_{t,i}}})^2
\end{dmath}
We can reorganize the above equation as,
\begin{dmath}
g_{t,i}(\theta_{t,i}-\theta_{,i}^*)=\frac{(1-\beta_1^t)\sqrt{\hat{v}_{t,i}}}{\alpha_t(1-\beta_{1,t})}
\Big((\theta_{t,i}-\theta_{,i}^*)^2-(\theta_{t+1,i}-\theta_{,i}^*)^2\Big)
+\frac{2\beta_{1,t}}{1-\beta_{1,t}}(\theta_{,i}^*-\theta_{t,i})s_{t-1,i}
+\frac{\alpha_t(1-\beta_1^t)}{(1-\beta_{1,t})}\frac{(\hat{s}_{t,i})^2}{\sqrt{\hat{v}_{t,i}}}
- \Delta \theta g_{t,i}^2(\theta_{t,i}-\theta_{,i}^*).
\end{dmath}
We can rewrite it as follows:
\begin{dmath}
g_{t,i}(\theta_{t,i}-\theta_{,i}^*)=\frac{(1-\beta_1^t)\sqrt{\hat{v}_{t,i}}}{\alpha_t(1-\beta_{1,t})}
\Big((\theta_{t,i}-\theta_{,i}^*)^2-(\theta_{t+1,i}-\theta_{,i}^*)^2\Big)
 +\sqrt{\frac{2\beta_{1,t}}{\alpha_{t-1}(1-\beta_{1,t})}(\theta_{,i}^*-\theta_{t,i})^2\sqrt{\hat{v}_{t-1,i}}} \sqrt{\frac{2\beta_{1,t}\alpha_{t-1}(s_{t-1,i})^2}{(1-\beta_{1,t})\sqrt{\hat{v}_{t-1,i}}}}
+\frac{\alpha_t(1-\beta_1^t)}{(1-\beta_{1,t})}\frac{(\hat{s}_{t,i})^2}{\sqrt{\hat{v}_{t,i}}}
- \Delta \theta g_{t,i}^2(\theta_{t,i}-\theta_{,i}^*)
\end{dmath}
Next, we use the Young's inequality, $ab \leq a^2/2+b^2/2$ as well as the information that $\beta_{1,t} \leq \beta_1$. We also replace $\Delta \theta$ with $\theta_{t-1} - \theta_{t}$ Thus, we can write the above equation as,
\begin{dmath}
g_{t,i}(\theta_{t,i}-\theta_{,i}^*) \leq \frac{1}{\alpha_t(1-\beta_1)}
\Big((\theta_{t,i}-\theta_{,i}^*)^2-(\theta_{t+1,i}-\theta_{,i}^*)^2\Big)\sqrt{\hat{v}_{t,i}}
 +\frac{\beta_{1,t}}{\alpha_{t-1}(1-\beta_{1,t})}(\theta_{,i}^*-\theta_{t,i})^2\sqrt{\hat{v}_{t-1,i}} 
 + \frac{\beta_1\alpha_{t-1}(s_{t-1,i})^2}{(1-\beta_1)\sqrt{\hat{v}_{t-1,i}}}
+\frac{\alpha_t}{(1-\beta_1)}\frac{(\hat{s}_{t,i})^2}{\sqrt{\hat{v}_{t,i}}}
- (\theta_{t-1,i}-\theta_{t,i})(\theta_{t,i}-\theta_{,i}^*)g_{t,i}^2
\end{dmath}
In order to compute the regret bound, we aggregate it as per the Lemma 10.4 of Aadm \cite{Adam} across the dimensions for $i\in \{1,\dots,d\}$ and the convex function sequence for $t\in \{1,\dots,T\}$ in the upper bound of $f_t(\theta_t)-f_t(\theta^*)$ as,
\begin{dmath}
R(T) \leq \sum_{i=1}^{d}{\frac{1}{\alpha_1(1-\beta_1)}} (\theta_{1,i}-\theta_{,i}^*)^2\sqrt{\hat{v}_{1,i}} 
+ \sum_{i=1}^{d}{\sum_{t=2}^{T}{\frac{1}{(1-\beta_1)}} (\theta_{t,i}-\theta_{,i}^*)^2(\frac{\sqrt{\hat{v}_{t,i}}}{\alpha_t}-\frac{\sqrt{\hat{v}_{t-1,i}}}{\alpha_{t-1}})}
  + \sum_{i=1}^{d}{\sum_{t=1}^{T}{\frac{\beta_{1,t}}{\alpha_{t}(1-\beta_{1,t})}(\theta_{,i}^*-\theta_{t,i})^2\sqrt{\hat{v}_{t,i}}}}
  + \frac{2\beta_1\alpha G_\infty}{(1-\beta_1)\sqrt{1-\beta_2}(1-\gamma)^2}\sum_{i=1}^{d}{||g_{1:T,i}||_2}
+ \frac{2\alpha G_\infty}{(1-\beta_1)\sqrt{1-\beta_2}(1-\gamma)^2}\sum_{i=1}^{d}{||g_{1:T,i}||_2}
+ \sum_{i=1}^{d}{\sum_{t=1}^{T}}(\theta_{t,i}-\theta_{t-1,i})(\theta_{t,i}-\theta_{,i}^*)g_{t,i}^2
\end{dmath}
It can be further refined with the assumptions that $\alpha=\alpha_t\sqrt{t}$, $||\theta_t-\theta^*||_2 \leq D$, $||\theta_m-\theta_n||_{\infty} \leq D_{\infty}$ and $\Delta \theta = (\theta_{t}-\theta_{t-1})$ is very small. Moreover, $\Delta \theta \approx 0$  as step size is very small when $t$ is large. Then, we can approximate $\Delta \theta$ with an upper bound of $1/t^2$. Thus, the above equation can be written as,
\begin{dmath}
R(T) \leq \frac{D^2}{\alpha(1-\beta_1)}\sum_{i=1}^{d}{\sqrt{T\hat{v}_{T,i}}} 
+ \frac{2\alpha(1+\beta_1) G_\infty}{(1-\beta_1)\sqrt{1-\beta_2}(1-\gamma)^2}\sum_{i=1}^{d}{||g_{1:T,i}||_2}
+ \frac{D_{\infty}^{2}}{\alpha}\sum_{i=1}^{d}{\sum_{t=1}^{t}{\frac{\beta_{1,t}}{(1-\beta_{1,t})}\sqrt{t\hat{v}_{t,i}}}}
+ D_\infty \sum_{i=1}^{d}{\sum_{t=1}^{T}}{\frac{\sqrt{t}}{t^2} g_{t,i}^2}
  \leq \frac{D^2}{\alpha(1-\beta_1)}\sum_{i=1}^{d}{\sqrt{T\hat{v}_{T,i}}} 
+ \frac{2\alpha(1+\beta_1) G_\infty}{(1-\beta_1)\sqrt{1-\beta_2}(1-\gamma)^2}\sum_{i=1}^{d}{||g_{1:T,i}||_2} 
+ \frac{D_{\infty}^{2}G_{\infty}\sqrt{1-\beta_2}}{\alpha}\sum_{i=1}^{d}{\sum_{t=1}^{t}{\frac{\beta_{1,t}}{(1-\beta_{1,t})}\sqrt{t}}}
+ D_\infty \sum_{i=1}^{d}{\sum_{t=1}^{T}}{\frac{g_{t,i}^2}{t}}
\end{dmath}
As per the finding of Adam \cite{Adam}, i.e., $\sum_{t=1}^{t}{\frac{\beta_{1,t}}{(1-\beta_{1,t})}\sqrt{t}} \leq \frac{1}{(1-\beta_1)(1-\gamma)^2}$, the regret bound can be further rewritten as,
\begin{dmath}
R(T) \leq \frac{D^2}{\alpha(1-\beta_1)}\sum_{i=1}^{d}{\sqrt{T\hat{v}_{T,i}}} 
+ \frac{2\alpha(1+\beta_1) G_\infty}{(1-\beta_1)\sqrt{1-\beta_2}(1-\gamma)^2}\sum_{i=1}^{d}{||g_{1:T,i}||_2} 
+ \sum_{i=1}^{d}{\frac{D_{\infty}^{2}G_{\infty}\sqrt{1-\beta_2}}{\alpha (1-\beta_1)(1-\lambda)^2}}
+ D_\infty \sum_{i=1}^{d}{\sum_{t=1}^{T}}{\frac{g_{t,i}^2}{t}}
\end{dmath}
Finally, we utilize Lemma 10.3 of Adam \cite{Adam} to approximate the upper bound as $\sum_{t=1}^{T}{\frac{g_{t,i}^2}{t}} \leq 4G_\infty^2||g_{1:T,i}||_2^2$. Thus, the final regret bound is given as,
\begin{dmath}
R(T) \leq \frac{D^2}{\alpha(1-\beta_1)}\sum_{i=1}^{d}{\sqrt{T\hat{v}_{T,i}}} 
+ \frac{2\alpha(1+\beta_1) G_\infty}{(1-\beta_1)\sqrt{1-\beta_2}(1-\gamma)^2}\sum_{i=1}^{d}{||g_{1:T,i}||_2} 
+ \sum_{i=1}^{d}{\frac{D_{\infty}^{2}G_{\infty}\sqrt{1-\beta_2}}{\alpha (1-\beta_1)(1-\lambda)^2}}
+ 4D_\infty G_\infty^2 \sum_{i=1}^{d}{||g_{1:T,i}||_2^2}
\end{dmath}
\end{proof}
\end{theorem}

\subsection*{B. Algorithms}
This section provides the Algorithms for different optimization techniques, including diffGrad (Algorithm \ref{algo:diffgrad}), diffGradInject (Algorithm \ref{algo:diffgradinject}), Radam (Algorithm \ref{algo:radam}), RadamInject (Algorithm \ref{algo:radaminject}), AdaBelief (Algorithm \ref{algo:adabelief}) and AdaBeliefInject (Algorithm \ref{algo:adabeliefinject}).

\begin{algorithm}[!h]
\caption{diffGrad Optimizer}
\SetAlgoLined
\textbf{Initialize:} $\theta_{0},m_{0}\gets0,v_{0}\gets0,t\gets0$\\
\textbf{Hyperparameters:} $\alpha, \beta_1, \beta_2$\\
\textbf{While} $\theta_{t}$ not converged\\
    \hspace{0.45cm} $t \gets t+1$\\
    \hspace{0.45cm} $g_t \gets \nabla_{\theta} f_t(\theta_{t-1})$ \\
    \hspace{0.45cm} $\xi_t \gets 1/(1+e^{-|g_t - g_{t-1}|})$ \\
    \hspace{0.45cm} $m_t \gets \beta_1 \cdot m_{t-1} + (1-\beta_1) \cdot g_t$\\
    \hspace{0.45cm} $v_t \gets \beta_2 \cdot v_{t-1} + (1-\beta_2) \cdot g^2_t$\\
    \hspace{0.45cm} \textbf{Bias Correction}\\
    \hspace{0.9cm} $\widehat{m_t} \gets m_t/(1-\beta_1^t)$, $\widehat{v_t} \gets v_t/(1-\beta_2^t)$\\
    \hspace{0.45cm} \textbf{Update}\\
    \hspace{0.9cm} $\theta_t \gets \theta_{t-1} - \alpha \xi_t \widehat{m_t}/(\sqrt{\widehat{v_t}} + \epsilon)$
\label{algo:diffgrad}
\end{algorithm}

\begin{algorithm}[!t]
\caption{diffGradInject (diffGrad + AdaInject) Optimizer}
\SetAlgoLined
\textbf{Initialize:} $\theta_{0},s_{0}\gets0,v_{0}\gets0,t\gets0$\\
\textbf{Hyperparameters:} $\alpha, \beta_1, \beta_2, k$\\
\textbf{While} $\theta_{t}$ not converged\\
    \hspace{0.45cm} $t \gets t+1$\\
    \hspace{0.45cm} $g_t \gets \nabla_{\theta} f_t(\theta_{t-1})$ \\
    \hspace{0.45cm} $\xi_t \gets 1/(1+e^{-|g_t - g_{t-1}|})$ \\
    \hspace{0.45cm} \textbf{If} t = 1 \\
    \hspace{0.9cm} $s_t \gets \beta_1 \cdot s_{t-1} + (1-\beta_1) \cdot g_t$\\
    \hspace{0.45cm} \textbf{Else} \\
    \hspace{0.9cm} $\textcolor{blue}{\Delta \theta \gets \theta_{t-2} - \theta_{t-1}}$\\
    \hspace{0.9cm} $s_t \gets \beta_1 \cdot s_{t-1} + (1-\beta_1) \cdot \textcolor{blue}{(g_t + \Delta \theta \cdot g^2_t)/k}$\\
    \hspace{0.45cm} $v_t \gets \beta_2 \cdot v_{t-1} + (1-\beta_2) \cdot g^2_t$\\
    \hspace{0.45cm} \textbf{Bias Correction}\\
    \hspace{0.9cm} $\widehat{s_t} \gets s_t/(1-\beta_1^t)$, $\widehat{v_t} \gets v_t/(1-\beta_2^t)$\\
    \hspace{0.45cm} \textbf{Update}\\
    \hspace{0.9cm} $\theta_t \gets \theta_{t-1} - \alpha \xi_t \widehat{s_t}/(\sqrt{\widehat{v_t}} + \epsilon)$
\label{algo:diffgradinject}
\end{algorithm}

\begin{algorithm}[!t]
\caption{Radam Optimizer}
\SetAlgoLined
\textbf{Initialize:} $\theta_{0},m_{0}\gets0,v_{0}\gets0,t\gets0$\\
\textbf{Hyperparameters:} $\alpha, \beta_1, \beta_2$\\
\textbf{While} $\theta_{t}$ not converged\\
    \hspace{0.45cm} $t \gets t+1$\\
    \hspace{0.45cm} $\rho_\infty \gets 2 / (1 - \beta_2) - 1$ \\
    \hspace{0.45cm} $g_t \gets \nabla_{\theta} f_t(\theta_{t-1})$ \\
    \hspace{0.45cm} $m_t \gets \beta_1 \cdot m_{t-1} + (1-\beta_1) \cdot g_t$\\
    \hspace{0.45cm} $v_t \gets \beta_2 \cdot v_{t-1} + (1-\beta_2) \cdot g_t^2$\\
    \hspace{0.45cm} $\rho_t = \rho_\infty - 2t\beta_2^t/(1-\beta_2^t)$\\
    \hspace{0.45cm} \textbf{If} $\rho_t \geq 5$\\
    \hspace{0.9cm} $\rho_u = (\rho_t - 4) \times (\rho_t - 2) \times \rho_\infty$\\
    \hspace{0.9cm} $\rho_d = (\rho_\infty - 4) \times (\rho_\infty - 2) \times \rho_t$\\
    \hspace{0.9cm} $\rho = \sqrt{(1 - \beta_2) \times \rho_u / \rho_d}$\\
    \hspace{0.9cm} $\alpha_1 = \rho \times \alpha / (1 - \beta_1^t)$\\
    \hspace{0.9cm} \textbf{Update}\\
    \hspace{1.35cm} $\theta_t \gets \theta_{t-1} - \alpha_1 \times m_t/(\sqrt{v_t} + \epsilon)$ \\
    \hspace{0.45cm} \textbf{Else}\\
    \hspace{0.9cm} $\alpha_2 = \alpha / (1 - \beta_1^t)$\\
    \hspace{0.9cm} \textbf{Update}\\
    \hspace{1.35cm} $\theta_t \gets \theta_{t-1} - \alpha_2 \times m_t$
\label{algo:radam}
\end{algorithm}

\begin{algorithm}[!t]
\caption{RadamInject (i.e., Radam + Inject) Optimizer}
\SetAlgoLined
\textbf{Initialize:} $\theta_{0},s_{0}\gets0,v_{0}\gets0,t\gets0$\\
\textbf{Hyperparameters:} $\alpha, \beta_1, \beta_2, k$\\
\textbf{While} $\theta_{t}$ not converged\\
    \hspace{0.45cm} $t \gets t+1$\\
    \hspace{0.45cm} $\rho_\infty \gets 2 / (1 - \beta_2) - 1$ \\
    \hspace{0.45cm} $g_t \gets \nabla_{\theta} f_t(\theta_{t-1})$ \\
    \hspace{0.45cm} \textbf{If} $t = 1$\\
    \hspace{0.9cm} $s_t \gets \beta_1 \cdot s_{t-1} + (1-\beta_1) \cdot g_t$\\
    \hspace{0.45cm} \textbf{Else}\\
    \hspace{0.9cm} \textcolor{blue}{$\Delta \theta \gets \theta_{t-1} - \theta_{t-2}$}\\
    \hspace{0.9cm} $s_t \gets \beta_1 \cdot s_{t-1} + (1-\beta_1) \cdot \textcolor{blue}{(g_t - \Delta \theta \cdot g_t^2)/k}$\\
    \hspace{0.45cm} $v_t \gets \beta_2 \cdot v_{t-1} + (1-\beta_2) \cdot g_t^2$\\
    \hspace{0.45cm} $\rho_t = \rho_\infty - 2t\beta_2^t/(1-\beta_2^t)$\\
    \hspace{0.45cm} \textbf{If} $\rho_t \geq 5$\\
    \hspace{0.9cm} $\rho_u = (\rho_t - 4) \times (\rho_t - 2) \times \rho_\infty$\\
    \hspace{0.9cm} $\rho_d = (\rho_\infty - 4) \times (\rho_\infty - 2) \times \rho_t$\\
    \hspace{0.9cm} $\rho = \sqrt{(1 - \beta_2) \times \rho_u / \rho_d}$\\
    \hspace{0.9cm} $\alpha_1 = \rho \times \alpha / (1 - \beta_1^t)$\\
    \hspace{0.9cm} \textbf{Update}\\
    \hspace{1.35cm} $\theta_t \gets \theta_{t-1} - \alpha_1 \times s_t/(\sqrt{v_t} + \epsilon)$ \\
    \hspace{0.45cm} \textbf{Else}\\
    \hspace{0.9cm} $\alpha_2 = \alpha / (1 - \beta_1^t)$\\
    \hspace{0.9cm} \textbf{Update}\\
    \hspace{1.35cm} $\theta_t \gets \theta_{t-1} - \alpha_2 \times s_t$
\label{algo:radaminject}
\end{algorithm}

\begin{algorithm}[!t]
\caption{AdaBelief Optimizer}
\SetAlgoLined
\textbf{Initialize:} $\theta_{0},m_{0}\gets0,v_{0}\gets0,t\gets0$\\
\textbf{Hyperparameters:} $\alpha, \beta_1, \beta_2$\\
\textbf{While} $\theta_{t}$ not converged\\
    \hspace{0.45cm} $t \gets t+1$\\
    \hspace{0.45cm} $g_t \gets \nabla_{\theta} f_t(\theta_{t-1})$ \\
    \hspace{0.45cm} $m_t \gets \beta_1 \cdot m_{t-1} + (1-\beta_1) \cdot g_t$\\
    \hspace{0.45cm} $v_t \gets \beta_2 \cdot v_{t-1} + (1-\beta_2) \cdot \textcolor{blue}{(g_t - m_t)^2}$\\
    \hspace{0.45cm} \textbf{Bias Correction}\\
    \hspace{0.9cm} $\widehat{m_t} \gets m_t/(1-\beta_1^t)$, $\widehat{v_t} \gets v_t/(1-\beta_2^t)$\\
    \hspace{0.45cm} \textbf{Update}\\
    \hspace{0.9cm} $\theta_t \gets \theta_{t-1} - \alpha \widehat{m_t}/(\sqrt{\widehat{v_t}} + \epsilon)$
\label{algo:adabelief}    
\end{algorithm}

\begin{algorithm}[!t]
\caption{AdaBeliefInject (AdaBelief + AdaInject) Optimizer}
\SetAlgoLined
\textbf{Initialize:} $\theta_{0},s_{0}\gets0,v_{0}\gets0,t\gets0$\\
\textbf{Hyperparameters:} $\alpha, \beta_1, \beta_2, k$\\
\textbf{While} $\theta_{t}$ not converged\\
    \hspace{0.45cm} $t \gets t+1$\\
    \hspace{0.45cm} $g_t \gets \nabla_{\theta} f_t(\theta_{t-1})$ \\
    \hspace{0.45cm} \textbf{If} t = 1 \\
    \hspace{0.9cm} $s_t \gets \beta_1 \cdot s_{t-1} + (1-\beta_1) \cdot g_t$\\
    \hspace{0.45cm} \textbf{Else} \\
    \hspace{0.9cm} $\textcolor{blue}{\Delta \theta \gets \theta_{t-2} - \theta_{t-1}}$\\
    \hspace{0.9cm} $s_t \gets \beta_1 \cdot s_{t-1} + (1-\beta_1) \cdot \textcolor{blue}{(g_t + \Delta \theta \cdot g^2_t)/k}$\\
    \hspace{0.45cm} $v_t \gets \beta_2 \cdot v_{t-1} + (1-\beta_2) \cdot \textcolor{blue}{(g_t - s_t)^2}$\\
    \hspace{0.45cm} \textbf{Bias Correction}\\
    \hspace{0.9cm} $\widehat{s_t} \gets s_t/(1-\beta_1^t)$, $\widehat{v_t} \gets v_t/(1-\beta_2^t)$\\
    \hspace{0.45cm} \textbf{Update}\\
    \hspace{0.9cm} $\theta_t \gets \theta_{t-1} - \alpha \widehat{s_t}/(\sqrt{\widehat{v_t}} + \epsilon)$
\label{algo:adabeliefinject}
\end{algorithm}

\end{document}